\documentclass[10pt,twocolumn,letterpaper]{article}

\usepackage[pagenumbers]{cvpr} % To force page numbers, e.g. for an arXiv version

\usepackage{multirow}
\usepackage{ifthen}
\usepackage{graphicx}
\usepackage{algorithm}
\usepackage{algorithmic}
\usepackage{comment}
\usepackage[table]{xcolor}

\usepackage{xcolor}
\usepackage{amsmath}
\usepackage{amssymb}
\usepackage[table]{xcolor}
\usepackage{threeparttable} % 表注必备
\usepackage[normalem]{ulem}
\usepackage{wrapfig}
\usepackage{adjustbox}
\usepackage[table]{xcolor}

\usepackage{xcolor}

\newboolean{K_is_15}

\setboolean{K_is_15}{false}

\definecolor{fftcolor}{RGB}{210, 10, 0}     % Professional Blue for FFT
\definecolor{loracolor}{RGB}{0, 80, 180}    % Professional Amber/Orange for LoRA

\definecolor{cvprblue}{rgb}{0.21,0.49,0.74}
\usepackage[pagebackref,breaklinks,colorlinks,allcolors=cvprblue]{hyperref}

\def\paperID{XXX} % *** Enter the Paper ID here
\def\confName{CVPR}
\def\confYear{2027}

\title{Beyond Uniform Subspaces: Spectrum-Aware and Depth-Adaptive Fusion for Multi-Task Model Merging}

\author{
Ruxi Gu$^{1,2}$ \quad Zilei Wang$^{1*}$ \quad Wei Wang$^{2,3*}$\\
$^{1}$Department of Automation, University of Science and Technology of China\\
$^{2}$State Key Laboratory of General Artificial Intelligence, BIGAI\\
$^{3}$School of Computer and Communication Engineering, University of Science and Technology Beijing\\
{\tt\small guruxi@mail.ustc.edu.cn, zlwang@ustc.edu.cn, wangwei@ustb.edu.cn}\\
$^{*}$Corresponding authors
}

\begin{document}
\maketitle
\begin{abstract}
Model merging aims to consolidate multiple task-specific models without access to extra training process. However, existing subspace-based methods largely rely on a uniform treatment of task updates, overlooking their intrinsic spectral and depth-wise heterogeneity. We identify two key deviations from this assumption: different tasks require different subspace capacity and exhibit different tolerance to spectral transformation, while subspace projection introduces depth-dependent distortion. Based on these observations, we propose \textbf{SADA-Merging}, a spectrum-aware and depth-adaptive framework for data-free model merging. SADA-Merging allocates task-specific subspace capacity according to spectral complexity, adapts spectral preservation according to task-wise plasticity, and applies depth-dependent anchoring to compensate for projection-induced distortion. This enables the fusion process to adapt to both the intrinsic geometry of each task and its sensitivity across network depth. SADA-Merging operates directly on task updates and is applicable to both full fine-tuning and LoRA settings. Extensive experiments demonstrate consistent improvements over existing data-free merging methods across different task scales and adaptation settings.
\end{abstract}    

\section{Introduction}
\label{sec:intro}

Pre-trained models provide a common foundation for adapting to diverse downstream tasks~\cite{radford2021learning, zhai2023sigmoid, wortsman2022robust, DINO}. Although independently fine-tuning a model for each task is effective, storing and deploying multiple task-specific models becomes increasingly costly. Model merging offers an attractive alternative by consolidating multiple task adaptations into a single model without joint retraining~\cite{ilharco2023task, yadav2023tiesmerging, li2023deepmodelfusionsurvey}. Existing approaches alleviate task interference through sign consensus~\cite{yadav2023tiesmerging}, task-vector alignment~\cite{DaheimMPGK24}, parameter selection and rescaling~\cite{MarczakTTC24}, and subspace-based fusion~\cite{tsv, cheng2025whoever}. Despite these advances, merging an increasing number of heterogeneous tasks remains challenging, particularly in the data-free setting.

Subspace-based methods provide a natural way to reduce interference by representing task updates in low-dimensional spaces~\cite{tsv, marczak2025notaskleftbehind, li2026essential}. However, existing approaches largely rely on a uniform treatment of task subspaces. This assumption overlooks two fundamental properties of task updates. First, tasks exhibit different spectral complexity: some concentrate their adaptation in a few dominant directions, whereas others distribute it across a broader spectrum. Thus, different tasks require different amounts of subspace capacity. Moreover, spectral concentration also reflects how sensitive a task is to spectral transformation: concentrated updates are more sensitive to perturbation, whereas spectrally distributed updates can tolerate greater modification. These observations motivate a task-adaptive fusion strategy that determines \emph{how much capacity} a task receives and \emph{how much transformation} its update undergoes based on the task vectors themselves.

Task heterogeneity, however, is not the only source of non-uniformity. We further observe that projecting task updates into a shared subspace introduces depth-dependent distortion: the amount of task-specific information preserved by the projection varies across network layers. Consequently, a uniform fusion strength across depth can be suboptimal, motivating depth-conditioned preservation in addition to task-conditioned adaptation.

Based on these observations, we propose \textbf{SADA-Merging} (\textbf{S}pectrum-\textbf{A}ware and \textbf{D}epth-\textbf{A}daptive Merging), a fully data-free framework that moves beyond uniform subspace fusion. SADA-Merging allocates task-specific subspace capacity according to spectral entropy, estimates task-wise plasticity from spectral concentration to control spectral transformation, and introduces depth-dependent anchoring to compensate for projection-induced distortion. Together, these components adapt the fusion process to both the intrinsic spectral structure of each task and its sensitivity across network depth.

We evaluate SADA-Merging on ViT models under both full fine-tuning (FFT) and low-rank adaptation (LoRA) \cite{hu2021lora} settings with varying numbers of tasks. We further assess its generalization to unseen tasks, where the merging strategy is evaluated on task configurations not encountered during the merging process. SADA-Merging consistently outperforms strong data-free merging baselines across both seen and unseen task settings, demonstrating the effectiveness of explicitly modeling task-wise spectral heterogeneity and depth-wise fusion sensitivity.

Our contributions are summarized as follows:
\begin{itemize}
    \item We identify the limitation of uniform subspace treatment in  model merging, and reveal two sources of non-uniformity: \emph{task-wise spectral heterogeneity} and \emph{depth-wise fusion sensitivity}.
    \item We introduce spectrum-aware task adaptation, where spectral entropy determines task-specific subspace capacity and spectral concentration determines the degree of spectral transformation.
    \item We propose depth-dependent anchoring to compensate for layer-dependent distortion introduced by shared-subspace projection.
    \item We develop \textbf{SADA-Merging}, a fully data-free framework applicable to both FFT and LoRA task updates, achieving consistent improvements across ViT models and different task scales, with strong generalization to unseen tasks.
\end{itemize}
\section{Related Works}

%\subsection{Model Merging}

Model merging seeks to combine independently adapted models into a single model while retaining their task capabilities without joint retraining. A common formulation is to represent each adaptation -- either a full fine-tuning update or a parameter-efficient LoRA update -- as a task vector relative to a shared pretrained model, after which the task vectors are combined to obtain the merged update~\cite{ilharco2023task}. Early work primarily focused on how to compose these updates in parameter space. TIES~\cite{yadav2023tiesmerging} reduces destructive interference through sign-based parameter selection, while subsequent approaches introduce magnitude-aware sparsification~\cite{yu2024language}, adaptive task coefficients~\cite{yangadamerging}, and task-vector alignment or functional transformations~\cite{DaheimMPGK24,cheng2025whoever,shi2025modelmergingfunctionaldual}. These methods establish task vectors as a convenient interface for data-free model composition, but largely treat the parameter space itself as the fusion space.

A different perspective has emerged from observing that task updates are often highly structured and approximately low-rank. Rather than deciding only \emph{which parameters} to retain, subspace-based approaches ask \emph{which directions} should participate in the merge. TSV-M~\cite{tsv} uses singular-value decomposition to expose task-specific low-dimensional structure, while Iso-CTS~\cite{marczak2025notaskleftbehind} further separates shared and task-specific components and normalizes their spectral scales. More recent work explores how a common coordinate system can be constructed across task updates. SCORE~\cite{Chaves_2026_CVPR}, for example, builds a shared orthogonal basis from leading singular directions and suppresses conflicting off-diagonal components, targeting subspace conflicts arising under domain shifts. ESM~\cite{li2026modelmergingessentialsubspace} takes a complementary functional perspective by identifying an essential subspace from feature shifts induced by the task updates, while PAVE~\cite{an2025purifying} uses task data to construct a knowledge-aware subspace and perform spectrum-based rank allocation. These developments indicate that the geometry of the representation space, rather than parameter-wise combination alone, is becoming central to model merging. 

Beyond static data-free merging, WEMoE~\cite{shen2025efficient} introduces input-conditioned routing, while TAP~\cite{de2026task} uses unlabeled data as a task-agnostic proxy for hyperparameter selection. For parameter-efficient merging, RobustMerge~\cite{zeng2026robustmerge} and NSC~\cite{lee2026label} specifically exploit the structure of LoRA updates through directional robustness and null-space compression, respectively. These approaches have demonstrated the benefits of adaptive routing, data-assisted selection, and update-specific compression for model merging.

Despite this progress, existing subspace methods predominantly focus on constructing or selecting an appropriate fusion space, while leaving the treatment of individual task updates relatively homogeneous once the space is determined. Although some recent approaches recognize task-specific heterogeneity, they rely on additional task-related data to estimate functional importance or guide subspace selection, introducing dependencies beyond the task vectors themselves. More fundamentally, task vectors can exhibit substantially different spectral structures even within the same model and layer. A concentrated update and a spectrally diffuse update should not necessarily receive the same representation capacity, nor should they be subjected to the same degree of spectral transformation. % Moreover, the effect of projecting an update into a shared space can vary across network depth, introducing another source of non-uniformity that is independent of task identity. 

Therefore, SADA-Merging takes a different view of subspace fusion. Rather than seeking a single fusion space for all task updates, we adapt how each task is allocated, transformed, and preserved within the shared space.

\section{Method}
\label{sec:method}

\subsection{Preliminaries}
\label{sec:preliminaries}

Given a pre-trained model with parameters $W_0$, we consider $N$ task-specific models $\{W^{(i)}\}_{i=1}^{N}$ obtained by independently adapting $W_0$ to different downstream tasks. Following Task Arithmetic~\cite{ilharco2023task}, we represent the adaptation for task $i$ by its task vector,
\begin{equation}
\Delta W^{(i)} = W^{(i)}-W_0.
\end{equation}

The objective of model merging is to obtain a single model that retains the capabilities of multiple task-specific models while minimizing task interference. We consider the \emph{data-free} setting, in which merging relies solely on the task vectors ${\Delta W^{(i)}}_{i=1}^N$, without access to training or proxy data as in ESM~\cite{li2026modelmergingessentialsubspace}. Specifically, we combine the task-specific updates using a merging operator $\mathcal{M}$,
\begin{equation}
\Delta W_{\mathrm{merge}}
=
\mathcal{M}\left(
\{\Delta W^{(i)}\}_{i=1}^{N}
\right),
\end{equation}
yielding the merged model $W_{\mathrm{merge}}=W_0+\Delta W_{\mathrm{merge}}$.

For non-matrix parameters, such as biases, we directly average the corresponding task vectors, as they generally lack meaningful subspace structure and play a minor role in model merging. Our focus is on matrix-valued updates, where task-specific spectral and subspace structures are more consequential. For a matrix-valued parameter at layer $l$, we denote its task-specific update by $\Delta W_l^{(i)}$.

\subsection{Motivation}
\label{sec:motivation}

\begin{figure}[ht]
     \centering
     \includegraphics[width=0.8\linewidth]{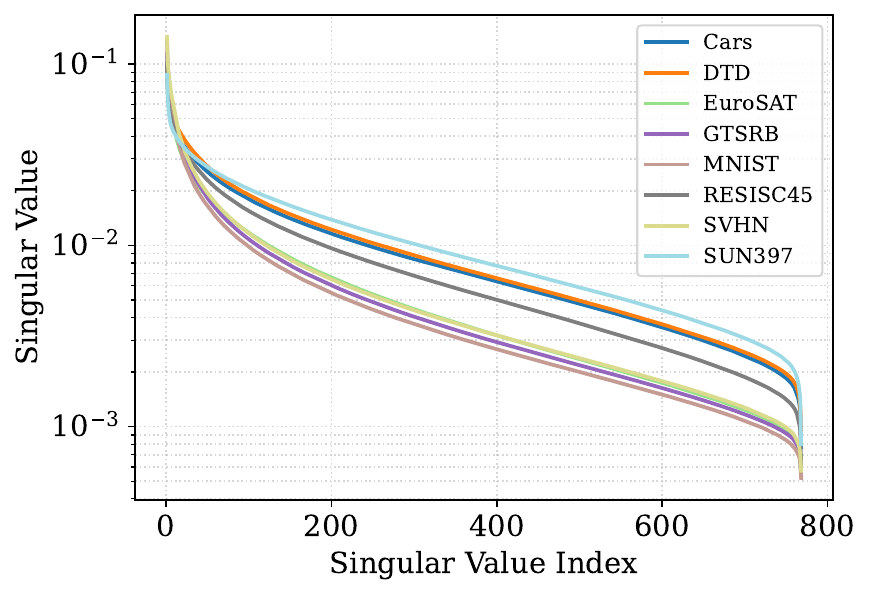}
     \caption{Singular value distributions of task vectors, averaged across all matrix-valued layers.}
     \label{fig:motivation1}
\end{figure}

\noindent{\textbf{Task-wise Spectral Heterogeneity.}}
Our first observation is that task updates exhibit markedly different spectral complexity. we perform singular value decomposition (SVD) on each task-specific update:
$\Delta W_l^{(i)}=U_{i,l}\Sigma_{i,l}V_{i,l}^\top$, where $\Sigma_{i,l}=\operatorname{diag}(\sigma_{i,l,1},\ldots,\sigma_{i,l,m_i})$ contains the singular values in descending order. The singular spectrum characterizes how the magnitude of a task update is distributed across orthogonal parameter directions. As shown in Fig.~\ref{fig:motivation1}, different tasks exhibit different spectral decay patterns. Relatively simple tasks, such as MNIST~\cite{MNIST} and SVHN~\cite{svhn}, show steep spectral decay, with most of their update magnitude concentrated in the leading components. In contrast, more complex tasks, such as SUN397~\cite{sun397}, DTD~\cite{dtd}, and Cars~\cite{cars}, retain markedly larger singular values over the same spectral indices, indicating that their updates span a broader set of meaningful directions.

This observation suggests that task updates do not require the same amount of subspace capacity. A uniform rank implicitly assumes that all tasks have comparable spectral complexity, which can unnecessarily allocate directions to highly concentrated updates while truncating useful directions from spectrally distributed ones.

\noindent{\textbf{Spectral Concentration and Task Plasticity.}}
The observed spectral heterogeneity provides not only a basis for allocating subspace capacity, but also a signal for determining how aggressively each task update can be transformed. An update whose spectral information is concentrated in a few leading directions relies more strongly on its original spectral structure and should therefore be preserved more carefully. Conversely, when the spectral magnitude is distributed across a broader set of directions, modifying its spectrum is less likely to alter a single dominant component that carries most of the task-specific information.

We quantify this property using the Essential Component Ratio (ECR), defined as the fraction of total singular-value mass contained in the leading $K$ components:
\begin{equation}
\operatorname{ECR}_{i,l}
=
\sum_{j=1}^{K}\sigma_{i.l,j}
\Big/
\sum_j\sigma_{i,l,j}.
\end{equation}
A larger ECR indicates that the update is more spectrally concentrated and hence more rigid to spectral modification. We map this concentration measure to a continuous Content-aware Plasticity Score (CPS):
\begin{equation}
\operatorname{CPS}_{i,l}
=
\operatorname{sigmoid}
\left(
\beta
\left(
\tau-\operatorname{ECR}_{i,l}
\right)
\right),
\end{equation}
where $\tau$ determines the transition point and $\beta$ controls the sharpness of the transition. Thus, a higher CPS corresponds to a more distributed, relatively plastic update and permits stronger spectral adaptation.

\begin{figure}[ht]
     \centering
     \includegraphics[width=0.8\linewidth]{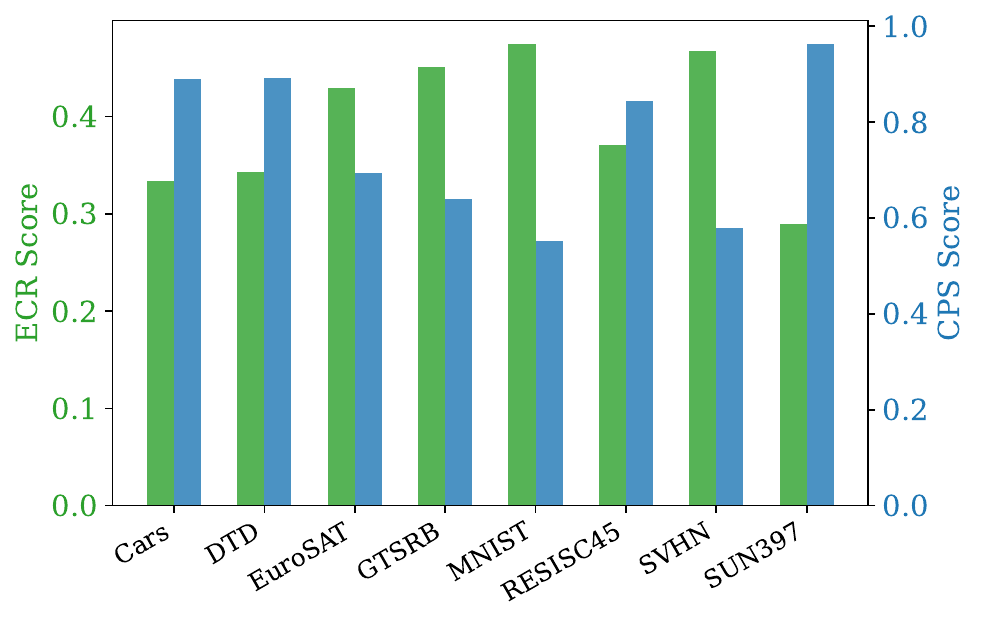}
     \caption{Comparison of Essential Component Ratio (ECR) and Content-aware Plasticity Score (CPS) across tasks.}
     \label{fig:motivation2}
\end{figure}

Fig.~\ref{fig:motivation2} provides evidence for this task-dependent distinction. MNIST and SVHN exhibit relatively high ECR and consequently low CPS, indicating that their spectral information is concentrated in a small number of dominant directions. In contrast, SUN397 and DTD exhibit lower ECR and higher CPS, reflecting their more distributed spectral structure. These results suggest that spectral concentration can serve as a data-free signal for determining how much spectral transformation each task can tolerate, rather than applying an identical transformation rule to all task updates.

\noindent{\textbf{Depth-dependent Fusion Sensitivity.}}
Task-adaptive spectral processing, however, does not eliminate another source of non-uniformity introduced by the fusion process itself. The effect of projecting task updates into a shared cover space can vary substantially across network depth. To examine this behavior, we visualize the \emph{spectral entropy} of the task updates at each layer. Spectral entropy characterizes how broadly the update magnitude is distributed across its singular directions. Given the singular values $\{\sigma_{i,l,j}\}$ of task $i$ at layer $l$, we first normalize them as
$
p_{i,l,j}
=
\frac{\sigma_{i,l,j}}
{\sum_k \sigma_{i,l,k}}
$, and compute the corresponding spectral entropy as
\begin{equation}
H_{i,l}
=
-\sum_j p_{i,l,j}\log p_{i,l,j}.
\label{eq5}
\end{equation}
A lower entropy indicates that the update is concentrated in a few dominant directions, whereas a higher entropy indicates a more distributed spectrum and thus greater spectral complexity.

\begin{figure}[ht]
     \centering
     \includegraphics[width=0.8\linewidth]{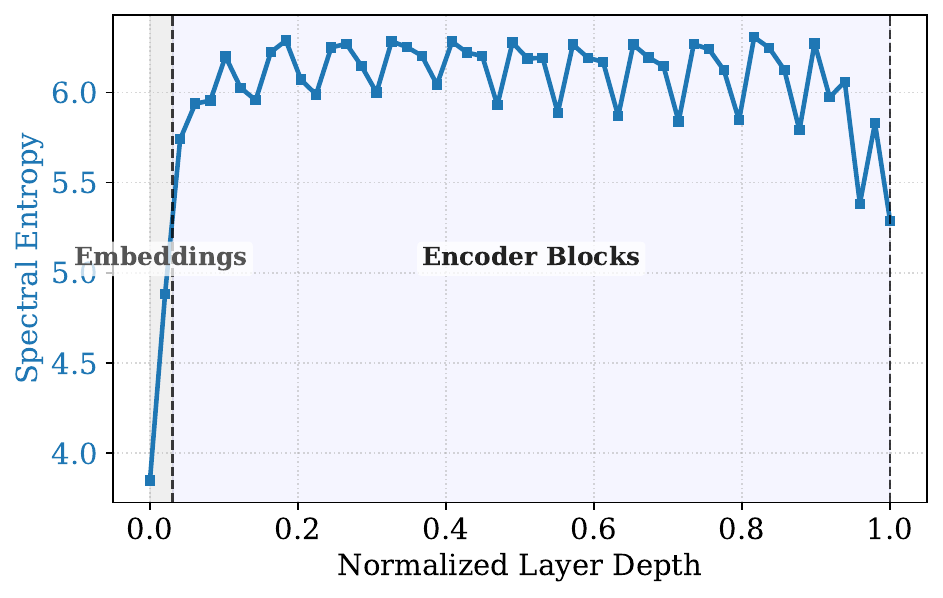}
     \caption{Depth-dependent characteristics of task updates under shared-space fusion.}
     \label{fig:motivation3}
\end{figure}

In Fig.~\ref{fig:motivation3}, within the encoder layers, the spectral entropy exhibits a consistent downward trend as the network goes deeper, revealing that the spectral geometry of task updates becomes progressively concentrated in deep layers. This decline in entropy indicates subspace energy shrinkage and implicit dimensional collapse toward the final layers, where task updates are squeezed into a narrow subspace of dominant singular vectors. Consequently, deep-layer updates become exceptionally vulnerable to projection-induced information loss during shared-space fusion. This layer-dependent spectral decay demonstrates that a static, layer-independent preservation scheme is inadequate.

\subsection{SADA-Merging}
\label{sec:sada}

Observations above reveal three distinct sources of heterogeneity that are overlooked by uniform subspace fusion. First, tasks differ in their spectral complexity and therefore require different representation capacities. Second, their spectral concentration differs, implying different tolerances to spectral modification. Third, the effect of shared-space projection varies across network depth, requiring layer-dependent preservation. These findings motivate SADA-Merging, which explicitly separates these three decisions: \emph{how much subspace capacity a task receives, how strongly its spectrum is allowed to change, and how strongly its task-specific component should be anchored during fusion}. SADA-Merging addresses them through elastic rank allocation, CPS-guided spectral adaptation, and task- and depth-dependent anchoring, respectively. The overall procedure is summarized in Algo.~\ref{alg:sada}, with implementation details provided in the supplementary material (Sec. \ref{A1}).

\begin{algorithm}[t]
\caption{SADA-Merging}
\label{alg:sada}
\begin{algorithmic}[1]
\REQUIRE Task vectors $\{\Delta W^{(i)}\}_{i=1}^{N}$
\ENSURE Merged task vector $\Delta W_{\mathrm{merge}}$
\FOR{each parameter layer $l$}
    \STATE $d_l\leftarrow l/(L-1)$
    \IF{$\Delta W_l^{(i)}$ is not a matrix}
        \STATE $\Delta W_l^{\mathrm{merge}}\leftarrow
        \frac{1}{N}\sum_{i=1}^{N}\Delta W_l^{(i)}$
        \STATE \textbf{continue}
    \ENDIF
    \FOR{$i=1,\ldots,N$}
        \STATE $(U_i,\Sigma_i,V_i^\top)
        \leftarrow\operatorname{SVD}(\Delta W_l^{(i)})$
        \STATE Compute $H_{i,l}$ and $\operatorname{CPS}_{i,l}$
        as in Sec.~\ref{sec:motivation}
    \ENDFOR
    \STATE Allocate $\{r_{i,l}\}_{i=1}^{N}$ from
    $\{H_{i,l}\}_{i=1}^{N}$ as in Sec.~\ref{sec:elastic}
    \FOR{$i=1,\ldots,N$}
        \STATE Retain $(U_i^r,\Sigma_i^r,V_i^{r\top})$
        according to $r_{i,l}$
        \STATE Construct $\widetilde{\Sigma}_i$ and
        $\Sigma_i^*$ according to Sec.~\ref{sec:cps}
        \STATE $\Delta W_{l,i}^*
        \leftarrow
        U_i^r\Sigma_i^*V_i^{r\top}$
        with CPS-guided adaptation
    \ENDFOR
    \STATE Construct $Q_U,Q_V$ and obtain $M_i$
    as in Sec.~\ref{sec:cover}
    \STATE Extract each task-specific diagonal block
    $M_{i,l}^{\mathrm{task}}$
    \FOR{$i=1,\ldots,N$}
        \STATE $a_{i,l}\leftarrow
        1+\lambda_{\mathrm{rigid}}(1-\operatorname{CPS}_{i,l})
        +\lambda_{\mathrm{depth}}d_l$
        \STATE $M_{\mathrm{merge},l}\leftarrow
        M_{\mathrm{merge},l}
        +a_{i,l}M_{i,l}^{\mathrm{task}}$
    \ENDFOR
    \STATE $\Delta W_l^{\mathrm{merge}}
    \leftarrow
    Q_{U,l}M_{\mathrm{merge},l}Q_{V,l}^{\top}$
    as in Sec.~\ref{sec:anchoring}
\ENDFOR
\RETURN $\Delta W_{\mathrm{merge}}$
\end{algorithmic}
\end{algorithm}

\subsubsection{Elastic Subspace Allocation}
\label{sec:elastic} For a decomposed task update
$\Delta W_l^{(i)}
=
U_{i,l}\Sigma_{i,l}V_{i,l}^{\top}$,
we use the spectral entropy $H_{i,l}$ introduced in Eq. \ref{eq5} to determine the effective rank. Specifically, within each layer, we standardize the task-wise entropy values and convert them into non-negative rank increments:
\begin{equation}
\delta r_{i,l}
=
\mathcal{R}(\hat{H}_{i,l}),
\end{equation}
where $\mathcal{R}(\cdot)$ maps the standardized entropy to a rank increment according to the available rank budget. The final rank $r_{i,l}$ is obtained by adding $\delta r_{i,l}$ to a layer-wise base rank $r_\mathrm{base}$ and enforcing the prescribed rank range. Thus, tasks with higher entropy receive larger subspaces, whereas spectrally concentrated tasks are represented with fewer directions. We retain the leading components:
$
U_{i,l}^{r}=U_{i,l}[:,1:r_{i,l}], 
\Sigma_{i,l}^{r}=\Sigma_{i,l}[1:r_{i,l},1:r_{i,l}], V_{i,l}^{r\top}=V_{i,l}^{\top}[1:r_{i,l},:].
$

\begin{figure}[ht]
     \centering
     \includegraphics[width=0.8\linewidth]{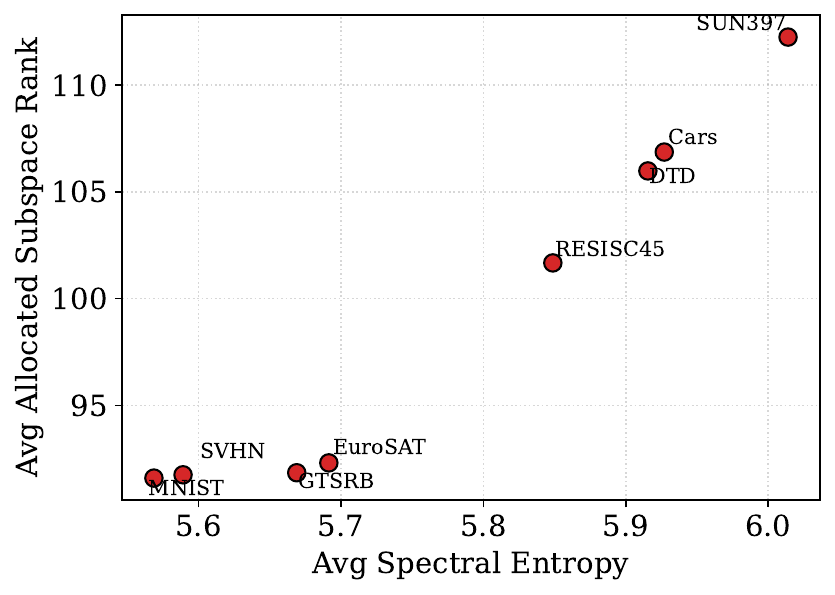}
     \caption{Relationship between spectral entropy and the allocated rank across tasks.}
     \label{fig:motivation4}
\end{figure}

As illustrated in Fig.~\ref{fig:motivation4}. This elastic allocation allows the representation capacity to follow the intrinsic spectral complexity of each task rather than imposing a uniform rank.

\subsubsection{CPS-guided Spectral Adaptation}
\label{sec:cps} The allocated rank determines how many spectral directions are retained, while CPS controls how strongly their spectral structure is adapted. For the retained singular values $\Sigma_{i,l}^{r}$, we construct a smoothed spectrum $\widetilde{\Sigma}_{i,l}$ by redistributing excessive spectral concentration toward lower-ranked components while preserving the overall update magnitude. We then use CPS to interpolate between the original and smoothed spectra:
\begin{equation}
\Sigma_{i,l}^{*}
=
(1-\operatorname{CPS}_{i,l})\Sigma_{i,l}^{r}
+
\operatorname{CPS}_{i,l}\widetilde{\Sigma}_{i,l},
\end{equation}
where $\Sigma_{i,l}^{r}$ preserves the task-specific spectral structure learned during adaptation, while $\widetilde{\Sigma}_{i,l}$ provides a more transferable spectral configuration by reducing excessive concentration. CPS therefore determines the balance between task-specific fidelity and spectral adaptability: a low CPS preserves more of the task's intrinsic structure, which is desirable for spectrally rigid tasks, whereas a high CPS allows greater spectral restructuring for tasks whose information is more distributed and consequently more amenable to adaptation during fusion. The resulting spectrally adapted update is
$
{\Delta W}_{l}^{*(i)}
=
U_{i,l}^{r}\Sigma_{i,l}^{*}V_{i,l}^{r\top}
$. Thus, spectrally rigid tasks with low CPS remain close to their original spectra, whereas more plastic tasks undergo stronger spectral smoothing before fusion.

\begin{table*}[t]
  \caption{Performance of model merging methods under the FFT setting. Main values denote average absolute accuracy, with average normalized accuracy reported in parentheses. \textbf{Bold} indicates the best result, while \underline{underlined} indicates the second-best. \emph{Zeroshot} denotes the original non-finetuned model, while \emph{Individual} denotes independently finetuned models. }
  \resizebox{1\textwidth}{!}{
    \begin{tabular}{cccc|ccc|ccc}
      \toprule
      \multicolumn{1}{c}{\multirow{2}{*}[-0.5ex]{\textbf{Method}}}         & \multicolumn{3}{c}{ViT-B-32}                     & \multicolumn{3}{c}{ViT-B-16}                     & \multicolumn{3}{c}{ViT-L-14}                    \\ 
                                               \cmidrule{2-10}
                                               & 8 tasks           & 14 tasks          & 20 tasks           & 8 tasks           & 14 tasks          & 20 tasks          & 8 tasks           & 14 tasks          & 20 tasks          \\ 
                                               \midrule
      \multicolumn{1}{c}{Zeroshot}            & $48.13$ & $57.12$ & $57.70$ & $55.51$ & $61.36$ & $62.10$ & $65.28$ & $68.53$ & $67.93$ \\
      \multicolumn{1}{c}{Individual}            & $92.82$ & $90.87$ & $91.36$ & $94.64$ & $92.75$ & $93.17$ & $95.81$ & $94.28$ & $94.72$ \\ \midrule
      \multicolumn{1}{c}{Weight Averaging}    & $66.34_{(72.13)}$ & $64.34_{(71.12)}$ & $61.04_{(67.53)}$ & $72.22_{(76.60)}$ & $69.46_{(74.82)}$ & $65.31_{(70.36)}$ & $79.56_{(83.15)}$ & $76.73_{(81.10)}$ & $71.60_{(75.60)}$ \\
      \multicolumn{1}{c}{Task Arithmetic}     & $70.79_{(76.55)}$ & $65.32_{(72.09)}$ & $60.52_{(66.79)}$ & $75.41_{(79.58)}$ & $70.52_{(75.89)}$ & $65.78_{(70.76)}$ & $84.93_{(88.65)}$ & $79.41_{(83.95)}$ & $74.01_{(78.07)}$ \\
      \multicolumn{1}{c}{TIES-Merging} & $75.09_{(81.08)}$ & $68.02_{(74.83)}$ & $63.38_{(69.90)}$ & $79.74_{(84.34)}$ & $73.22_{(78.73)}$ & $68.18_{(73.26)}$ & $86.88_{(90.69)}$ & $79.46_{(84.05)}$ & $75.71_{(79.80)}$ \\
      \multicolumn{1}{c}{Consensus TA}        & $75.03_{(80.84)}$ & $70.39_{(77.36)}$ & $65.43_{(71.98)}$ & $79.39_{(83.86)}$ & $74.39_{(79.92)}$ & $69.76_{(74.93)}$ & $86.34_{(90.08)}$ & $82.22_{(86.94)}$ & $79.00_{(83.22)}$ \\

      \multicolumn{1}{c}{WUDI-Merging}        & $86.47_{(93.04)}$ & $78.87_{(89.71)}$ & $69.90_{(76.71)}$ & $89.77_{(94.75)}$ & $82.95_{(89.20)}$ & $74.96_{(80.32)}$ & $93.97_{(98.03)}$ & $89.74_{(95.06)}$ & $84.59_{(89.15)}$ \\
      
      \multicolumn{1}{c}{TSV-M} & $85.86_{(92.31)}$ & $80.06_{(87.88)}$ & $77.07_{(84.29)}$ & $89.01_{(93.94)}$ & $84.58_{(91.01)}$ & $80.57_{(86.45)}$ & $92.98_{(96.98)}$ & $89.17_{(94.43)}$ & $87.72_{(92.50)}$ \\
      
      \multicolumn{1}{c}{Iso-CTS}        & ${86.11_{(92.73)}}$ & ${81.46_{(89.45)}}$ & ${78.18_{(85.55)}}$ & \underline{$90.84_{{(95.93)}}$} & ${86.10_{(92.55)}}$ & ${82.72_{(88.68)}}$ & $\mathbf{94.79_{(98.89)}}$ & $\underline{91.05_{(96.36)}}$ & ${90.15_{(94.99)}}$ \\
      
      \multicolumn{1}{c}{DC-Merge} & $\underline{86.97_{(93.47)}}$ & $\underline{82.38_{(90.43)}}$ & $\underline{80.46_{(88.10)}}$ & ${90.48_{(95.52)}}$ & $\underline{86.80_{(93.48)}}$ & $\underline{84.40_{(90.58)}}$ & ${94.06_{(98.14)}}$ & ${90.80_{(96.22)}}$ & $\underline{90.21_{(95.20)}}$ \\
      \bottomrule

      % \rowcolor{gray!15}\multicolumn{1}{c}{\textbf{Ours}} & $\mathbf{88.36_{(95.16)}}$ & $\mathbf{83.98_{(92.29)}}$ & $\mathbf{82.15_{(89.85)}}$ & $\mathbf{91.08_{(96.19)}}$ & $\mathbf{87.41_{(94.15)}}$ & $\mathbf{85.05_{(91.28)}}$ & $\underline{94.52_{(98.63)}}$ & $\mathbf{91.37_{(96.83)}}$ & $\mathbf{90.73_{(95.75)}}$ \\

      \rowcolor{gray!15}\multicolumn{1}{c}{\textbf{Ours}} & $\mathbf{88.36_{(95.15)}}$ & $\mathbf{83.89_{(92.15)}}$ & $\mathbf{82.30_{(90.06)}}$ & $\mathbf{91.11_{(96.21)}}$ & $\mathbf{87.41_{(94.13)}}$ & $\mathbf{85.10_{(91.36)}}$ & $\underline{94.54_{(98.64)}}$ & $\mathbf{91.46_{(96.93)}}$ & $\mathbf{90.85_{(95.87)}}$ \\

      \bottomrule
    \end{tabular}
  }
  \label{tab:FFT_task_acc}
  
\end{table*}

\begin{table*}[!t]
  \caption{Performance of model merging methods under the LoRA setting. Zeroshot performance is the same as that reported in Tab. \ref{tab:FFT_task_acc}.}
  \resizebox{1\textwidth}{!}{
    \begin{tabular}{cccc|ccc|ccc}
      \toprule
      \multicolumn{1}{c}{\multirow{2}{*}[-0.5ex]{\textbf{Method}}}         & \multicolumn{3}{c}{ViT-B-32}                     & \multicolumn{3}{c}{ViT-B-16}                     & \multicolumn{3}{c}{ViT-L-14}                    \\ 
                                               \cmidrule{2-10}
                                               & 8 tasks           & 12 tasks          & 16 tasks           & 8 tasks           & 12 tasks          & 16 tasks          & 8 tasks           & 12 tasks          & 16 tasks          \\ 
                                               \midrule
      \multicolumn{1}{c}{Individual}            & $87.82$ & $88.90$ & $87.50$ & $89.71$ & $90.76$ & $89.11$ & $92.36$ & $93.57$ & $92.11$ \\ \midrule
      \multicolumn{1}{c}{Task Arithmetic}     & $52.80_{(61.73)}$ & $60.76_{(69.12)}$ & $60.04_{(68.94)}$ & $57.70_{(65.30)}$ & $64.26_{(71.12)}$ & $62.40_{(70.16)}$ & $68.29_{(74.38)}$ & $73.69_{(78.84)}$ & $69.98_{(75.58)}$ \\
      \multicolumn{1}{c}{KnOTS-TIES}    & $55.93_{(65.03)}$ & $63.03_{(71.43)}$ & $61.78_{(70.78)}$ & $60.80_{(68.59)}$ & $66.35_{(73.38)}$ & $64.31_{(72.29)}$ & $73.61_{(79.92)}$ & $75.64_{(80.90)}$ & $72.19_{(77.98)}$ \\
      \multicolumn{1}{c}{WUDI-Merging}        & $55.25_{(64.38)}$ & $62.20_{(70.64)}$ & $61.24_{(70.27)}$ & $58.95_{(66.63)}$ & $65.29_{(72.29)}$ & $64.59_{(72.51)}$ & $69.78_{(75.91)}$ & $74.25_{(79.45)}$ & $71.79_{(77.59)}$ \\
      \multicolumn{1}{c}{TSV-M} & $58.91_{(68.25)}$ & $65.30_{(73.86)}$ & $63.51_{(72.72)}$ & $62.97_{(70.87)}$ & $68.92_{(76.06)}$ & $67.21_{(75.35)}$ & $76.52_{(83.00)}$ & $79.67_{(85.13)}$ & $74.37_{(80.31)}$ \\
      \multicolumn{1}{c}{Iso-CTS} & ${63.01_{(72.71)}}$ & ${66.28_{(75.01)}}$ & ${64.61_{(74.02)}}$ & ${69.06_{(77.38)}}$ & ${71.52_{(78.87)}}$ & ${69.88_{(78.22)}}$ & ${81.64_{(88.31)}}$ & ${81.35_{(86.87)}}$ & ${77.50_{(83.65)}}$ \\

      \multicolumn{1}{c}{DC-Merge} & $\underline{64.12_{(73.84)}}$ & $\underline{68.36_{(77.18)}}$ & $\underline{66.26_{(75.78)}}$ & $\underline{70.50_{(78.81)}}$ & $\underline{73.13_{(80.57)}}$ & $\underline{70.54_{(78.87)}}$ & $\underline{82.60_{(89.40)}}$ & $\underline{83.62_{(89.31)}}$ & $\underline{79.55_{(85.74)}}$ \\
      \bottomrule
        \rowcolor{gray!15}
            \multicolumn{1}{c}{\textbf{Ours}} & $\mathbf{65.18_{(74.95)}}$ & $\mathbf{69.01_{(77.84)}}$ & $\mathbf{66.59_{(76.11)}}$ & $\mathbf{71.17_{(81.05)}}$ & $\mathbf{73.55_{(81.05)}}$ & $\mathbf{70.85_{(79.25)}}$ & $\mathbf{83.03_{(89.87)}}$ & $\mathbf{83.77_{(89.47)}}$ & $\mathbf{79.70_{(85.88)}}$ \\
      \bottomrule
    \end{tabular}
  }

  \label{tab:LoRA_task_acc}
\end{table*}

\subsubsection{Shared Cover-Space Fusion}
\label{sec:cover} After task-specific spectral adaptation, the retained subspaces remain heterogeneous across tasks. Following the cover-space formulation of DC-Merge~\cite{zhang2026dc}, we construct a shared orthogonal space from the retained left and right singular subspaces:
$U_l=[U_{1,l}^{r},\ldots,U_{N,l}^{r}], V_l^{\top}
=
[V_{1,l}^{r\top};\ldots;V_{N,l}^{r\top}]$, 
and orthogonalize them to obtain the shared cover matrices $Q_{U,l}$ and $Q_{V,l}$. Each adapted task update is then expressed in this common coordinate system:
\begin{equation}
M_{i,l}
=
Q_{U,l}^{\top}
\Delta W_{l}^{*(i)}
Q_{V,l}.
\end{equation}

SADA-Merging further preserves the task-specific structure induced by the elastic rank allocation. Let
$
b_i=\sum_{j=1}^{i}r_{j,l}
$
denote the cumulative rank offsets. We retain the diagonal block corresponding to the retained subspace of each task:
\begin{equation}
M_{i,l}^{\mathrm{task}}
=
M_{i,l}
[b_{i-1}:b_i,\,
b_{i-1}:b_i].
\end{equation}
This block-wise construction preserves each task's contribution within the shared cover space while allowing different tasks to occupy subspaces of different dimensions. Unlike a uniform cover-space fusion, the resulting task contributions therefore remain explicitly coupled to their adaptive ranks, providing the basis for the task- and depth-dependent anchoring introduced next.

\begin{figure}[ht]
     \centering
     \includegraphics[width=0.8\linewidth]{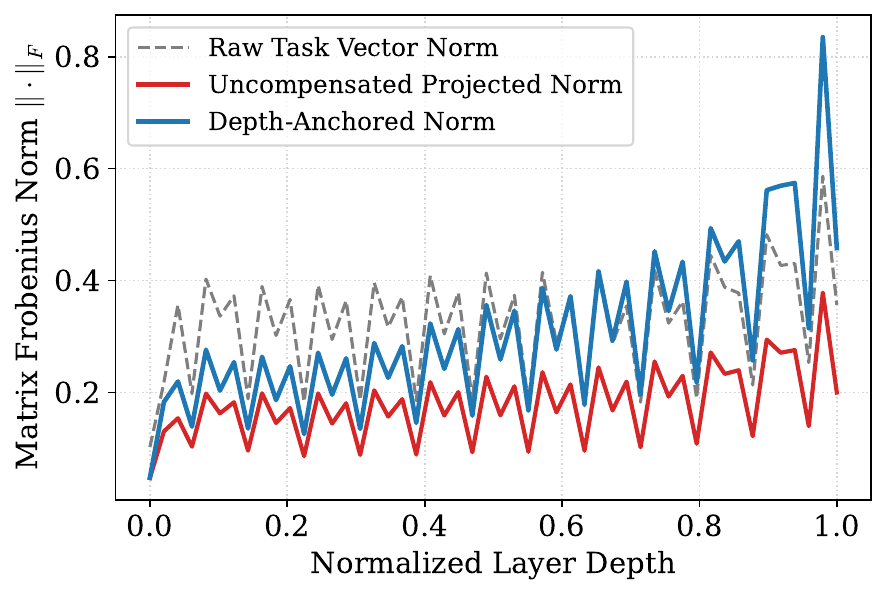}
     \caption{Depth-dependent attenuation of task-update magnitude and its compensation by depth anchoring.}
     \label{fig:motivation6}
\end{figure}

\subsubsection{Task- and Depth-dependent Anchoring}

\label{sec:anchoring} The spectral analysis in Fig.~\ref{fig:motivation3} shows that task-update spectra become progressively concentrated toward deeper layers, indicating increasing sensitivity to projection-induced information loss. We therefore introduce depth-aware compensation when aggregating task-specific components. For task $i$ at layer $l$, we use
\begin{equation}
a_{i,l}
=
1+
\lambda_{\mathrm{rigid}}
(1-\operatorname{CPS}_{i,l})
+
\lambda_{\mathrm{depth}}d_l,
\end{equation}
where $d_l$ is the normalized layer depth. The first term provides stronger preservation for spectrally rigid tasks, while the second progressively increases the contribution of deeper layers. Fig.~\ref{fig:motivation6} compares the original task-update norm with the projected norm before and after depth anchoring. The projected norm increasingly deviates from the original norm at deeper layers, while depth anchoring not only compensates for this attenuation but also reinforces the task-update magnitude toward deeper layers.

The merged cover-space representation is consequently structured as
$
M_{\mathrm{merge},l}
=
\sum_{i=1}^{N}
a_{i,l}M_{i,l}^{\mathrm{task}}
$, 
and the merged task update is reconstructed as
$
\Delta W_l^{\mathrm{merge}}
=
Q_{U,l}
M_{\mathrm{merge},l}
Q_{V,l}^{\top}
$.
This produces a fusion rule that jointly adapts to task-specific spectral rigidity and depth-dependent projection sensitivity.

\section{Experiments}

\begin{figure*}[htbp]
  \centering
    \includegraphics[width=0.8\linewidth]{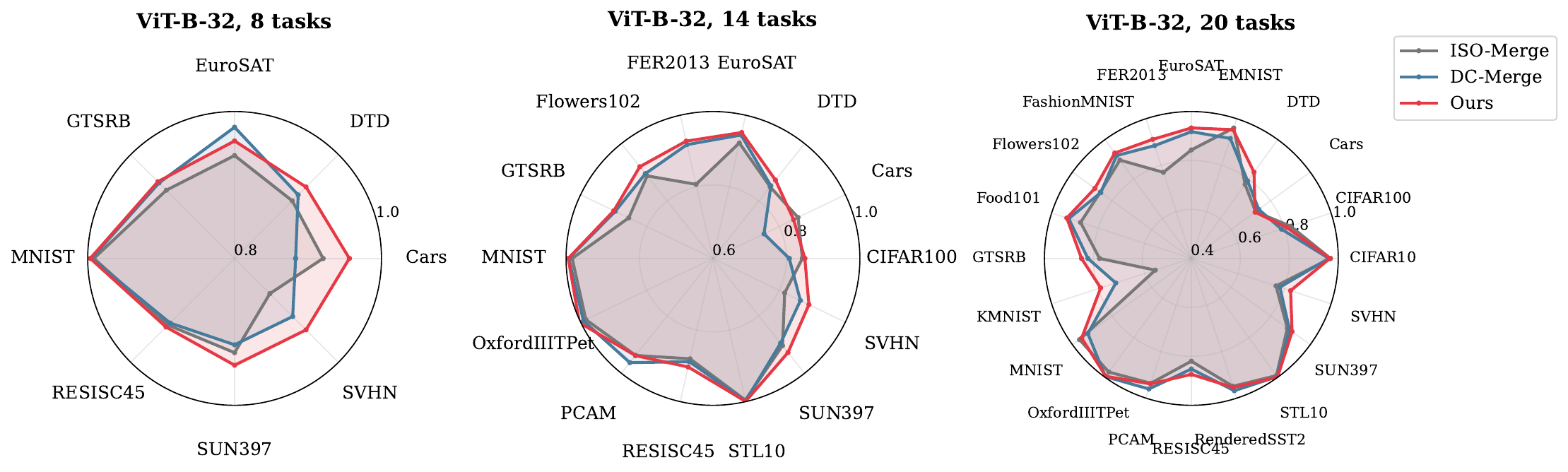}
    \label{fig:sub1}
    \caption{Radar charts of model merging performance on ViT-B-32 across 8, 14, and 20 tasks.}
    \label{fig6}
\end{figure*}

\subsection{Experimental Setup}
Following \cite{zhang2026dc}, we evaluate multi-task merging under three settings: (1) FFT merging on benchmarks with 8, 14, and 20 tasks; (2) LoRA merging on benchmarks with 8, 12, and 16 tasks; and (3) generalization to unseen tasks, where we merge task vectors from a smaller subset of tasks and evaluate the resulting model on additional tasks not included in the merge. For example, when merging 8 task vectors on the 14-task benchmark, the remaining 6 tasks are treated as unseen tasks. We use CLIP~\cite{radford2021learning} models with ViT-B-32, ViT-B-16, and ViT-L-14 visual encoders as pre-trained base models, and adopt the task-specific fine-tuned checkpoints provided by TALL-masks~\cite{wang2024localizing}. 

In the FFT setting, we compare SADA-Merging against Weight Averaging~\cite{wortsman2022model}, Task Arithmetic~\cite{ilharco2023task}, TIES-Merging~\cite{yadav2023tiesmerging}, Consensus TA~\cite{wang2024localizing}, WUDI-Merging~\cite{cheng2025whoever}, TSV-M~\cite{tsv}, Iso-CTS~\cite{marczak2025notaskleftbehind}, and DC-Merge~\cite{zhang2026dc}. In the LoRA setting, we compare against Task Arithmetic, KnOTS-TIES~\cite{stoica2024knots}, WUDI-Merging, TSV-M, Iso-CTS, and DC-Merge. For the unseen-task setting, we compare against the stronger baselines TSV-M, Iso-CTS, and DC-Merge.

For evaluation, we use the same protocol as \cite{zhang2026dc}, selecting the optimal scaling factor for $\Delta W_{\mathrm{merge}}$ on the validation set from 0.1 to 3.0 with a step size of 0.1, and then applying it to the test set. For both FFT and LoRA settings, we report the absolute accuracy and the normalized accuracy relative to the individually fine-tuned models. For unseen-task generalization, we report the absolute accuracy on the unseen tasks and on all test tasks. More details are provided in the supplementary material (Sec.~\ref{B1}).

\subsection{Experimental Results}

\noindent \textbf{Performance on the FFT setting.} As shown in Tab.~\ref{tab:FFT_task_acc}, SADA-Merging consistently achieves the best performance across different model architectures and numbers of merged tasks. On ViT-B-32, SADA-Merging improves over the previous best method, DC-Merge, by $1.39$, $1.51$, and $1.84$ percentage points for 8, 14, and 20 tasks, respectively.  Fig.~\ref{fig6} further shows that the gains are particularly pronounced for tasks with more extreme CPS values, such as SUN397 with high CPS and SVHN with low CPS, suggesting that explicitly modeling task-dependent spectral plasticity is especially beneficial when task updates exhibit markedly different degrees of spectral rigidity.

Similar improvements are observed on ViT-B-16, where SADA-Merging achieves $91.11$, $87.41$, and $85.10$ accuracy, consistently outperforming both Iso-CTS and DC-Merge. On the larger ViT-L-14, SADA-Merging remains competitive and achieves the best performance for 14 and 20 tasks, while obtaining the second-best result for 8 tasks. Overall, these results demonstrate that the proposed task- and depth-adaptive design provides consistent gains over existing merging methods in the FFT setting.

\noindent \textbf{Performance on the LoRA setting.} The results are reported in Tab.~\ref{tab:LoRA_task_acc}. SADA-Merging maintains a clear performance advantage across all three architectures and evaluated task scales, confirming that the proposed method remains effective for parameter-efficient task updates.

The improvements in the LoRA setting are generally smaller than those under FFT, due to the intrinsic spectral characteristics of LoRA updates. With the default $K=10\%$ for ECR computation, LoRA task vectors typically exhibit ECR values close to $1$, indicating that most essential spectral components are captured by the leading singular directions. Consequently, CPS values are mostly near $0$, weakly activating CPS-guided spectral adaptation and leaving less spectral heterogeneity for SADA-Merging to exploit. This observation is consistent with our motivation, that adaptive spectral transformation benefits from meaningful variation in task-update spectra. Detailed analysis are provided in the supplementary material (Sec. \ref{A3}).

\begin{table}[htbp]
  \caption{Performance of model merging on unseen tasks for the ViT-B-32 backbone. Each column follows the \texttt{test:N / train:M} notation, where \texttt{M} of the \texttt{N} tasks are used for merging and the remaining \texttt{N-M} tasks are unseen during merging. The first value indicates performance on unseen tasks, and the second indicates performance on all tasks.}
  \centering
  \resizebox{1.0\columnwidth}{!}{
    \begin{tabular}{l|ccc}
      \toprule      
      \textbf{Method}   &  test:14 / train:8        & test:20 / train:8     & test:20 / train:14 \\ \midrule
      Zeroshot        & $69.12/57.12$ & $64.08/57.70$ & $59.03/57.70$ \\ 
      Individual   & $88.28/90.87$ & $90.39/91.36$ & $92.51/91.36$  \\ \midrule
      
      TSV-M        & $64.22/75.96$ & $\underline{65.48}/72.72$ & $64.38/75.27$ \\ 
      Iso-CTS   & $63.58/76.21$ & $65.02/72.76$ & $64.03/76.23$  \\
      DC-Merge        & $\underline{64.73}/\underline{77.16}$ & $65.12/\underline{73.48}$ & $\underline{64.78}/\underline{77.16}$  \\ \midrule \rowcolor{gray!15}
      \textbf{Ours}     & $\mathbf{66.74/78.03}$ & $\mathbf{66.29/74.38}$ & $\mathbf{65.21/78.33}$  \\
      \bottomrule
    \end{tabular}
  }
  \label{tab:unseen_tasks}
\end{table}

\noindent \textbf{Performance on Unseen Tasks.} As shown in Tab.~\ref{tab:unseen_tasks}, SADA-Merging exhibits the strongest generalization to unseen tasks among the evaluated merging methods.

When only 8 tasks are available for merging, none of the merging methods surpasses Zeroshot on the 14-task benchmark, with SADA-Merging achieving $66.74$ compared with $69.12$ for Zeroshot. This suggests that a limited set of seen tasks provides insufficient coverage of the broader task distribution, making generalization beyond the observed tasks challenging. Nevertheless, SADA-Merging remains the strongest merging method.

On the fixed 20-task benchmark, increasing the number of seen tasks from 8 to 14 highlights the benefit of broader task coverage. With 8 seen tasks, SADA-Merging achieves $66.29$ on the 12 remaining unseen tasks and $74.38$ across all 20 tasks. With 14 seen tasks, it achieves $65.21$ on the remaining 6 unseen tasks and $78.33$ across all 20 tasks. Although the unseen-task sets differ between the two settings, the improvement in all-task accuracy indicates that incorporating more seen tasks allows SADA-Merging to better integrate a broader range of task-specific adaptations. SADA-Merging also maintains a clear advantage over competing merging methods in both settings, demonstrating robust generalization beyond the tasks used for merging. Results for the ViT-B-16 and ViT-L-14 backbones are provided in the supplementary material, where SADA-Merging also achieves better performance (Sec. \ref{B3}).

\subsection{Ablation Study and Sensitivity Analysis}

\noindent \textbf{Component Ablation.} All experiments in this section are conducted on ViT-B-32. We progressively introduce the three components of SADA-Merging to examine their individual contributions. The Base configuration performs shared-space fusion without CPS-guided spectral adaptation (CPS), elastic rank allocation (Rank), or depth-dependent anchoring (Depth). As shown in Tab.~\ref{tab:ablation1}, introducing CPS improves performance from $87.10$ to $87.37$ on 8 tasks and from $82.56$ to $83.00$ on 14 tasks, indicating that task-dependent spectral plasticity provides a consistent benefit. Adding elastic rank allocation further improves performance. This result supports our hypothesis that tasks with different spectral complexity should not be represented using a uniform subspace capacity.

\begin{table}[htbp]
  \caption{Ablation study of SADA-Merging with progressively added components.}
  \centering
  \resizebox{0.9\columnwidth}{!}{
    \begin{tabular}{l|ccc}
      \toprule      
      \textbf{Method}   & 8 tasks           & 14 tasks          & 20 tasks \\ 
      \midrule
      % Base        & $87.03$ & $82.49$ & $80.38$ \\ \midrule + CPS   & $87.42$ & $83.05$ & $80.86$  \\+ CPS + Rank        & $87.53$ & $83.21$ & $81.01$  \\+ CPS + Rank + Depth (Ours)     & $\mathbf{88.36}$ & $\mathbf{83.98}$ & $\mathbf{82.15}$  \\
      Base        & $87.10$ & $82.56$ & $80.54$ \\ \midrule + CPS   & $87.37$ & $83.00$ & $80.94$  \\+ CPS + Rank        & \underline{$87.53$} & \underline{$83.23$} & \underline{$81.15$}  \\+ CPS + Rank + Depth (Ours)     & $\mathbf{88.36}$ & $\mathbf{83.89}$ & $\mathbf{82.30}$  \\
      
      \bottomrule
    \end{tabular}
  }
  \label{tab:ablation1}
\end{table}

Adding depth-dependent anchoring yields the largest gain. This consistent improvement demonstrates the importance of accounting for depth-dependent projection sensitivity, with stronger anchoring helping preserve task-specific information in deeper layers. Overall, the complete method improves over Base by $1.33$, $1.49$, and $1.77$ points, highlighting the importance of depth-aware preservation with more heterogeneous task subspaces.

\noindent \textbf{Sensitivity to Depth Anchoring.}
% We evaluate the sensitivity to the depth anchoring factor $\lambda_{\mathrm{depth}}$ in Tab.~\ref{tab:ablation2}. SADA-Merging remains stable across a relatively broad range of values. Increasing $\lambda_{\mathrm{depth}}$ from $0.5$ to $2.5$ improves performance, showing that stronger depth-dependent preservation is beneficial. Performance then remains stable for larger values, with only minor variations across different task scales. The default value of $2.5$ achieves the best overall performance, suggesting that the proposed depth-aware mechanism does not rely on a narrowly tuned parameter.
We evaluate the sensitivity to the depth anchoring factor $\lambda_{\mathrm{depth}}$ in Tab.~\ref{tab:ablation2}. SADA-Merging remains robust across a relatively broad range of values, with performance improving as $\lambda_{\mathrm{depth}}$ increases from $0.5$ to $2.5$. This suggests that stronger depth-dependent preservation is beneficial up to a moderate level, whereas excessively large values can become overly restrictive and degrade performance. The default value of $2.5$ achieves the best overall performance, providing a favorable balance between depth-dependent preservation and subspace flexibility.

\begin{table}[htbp]
  \caption{Sensitivity analysis of the depth anchoring factor $\lambda_{\mathrm{depth}}$.}
  \centering
  \resizebox{0.8\columnwidth}{!}{
    \begin{tabular}{l|ccc}
      \toprule      
      \textbf{Method}   & 8 tasks           & 14 tasks          & 20 tasks \\ 
      \midrule
      % Ours ($\lambda_{\mathrm{depth}}=2.0$)        & $88.36$ & $\mathbf{83.98}$ & $\mathbf{82.15}$ \\ \midrule$\lambda_{\mathrm{depth}}=0.5$   & $87.98$ & $83.60$ & $81.48$  \\$\lambda_{\mathrm{depth}}=1.0$   & $88.22$ & $83.87$ & $81.91$  \\$\lambda_{\mathrm{depth}}=1.5$        & $88.28$ & $83.93$ & $82.02$  \\$\lambda_{\mathrm{depth}}=2.5$      & $\mathbf{88.38}$ & $83.96$ & $82.14$  \\
      Ours ($\lambda_{\mathrm{depth}}=2.5$)        & $\mathbf{88.36}$ & \underline{$83.89$} & $\mathbf{82.30}$ \\ \midrule$\lambda_{\mathrm{depth}}=0.5$   & $87.83$ & $83.48$ & $81.74$  \\$\lambda_{\mathrm{depth}}=1.5$   & \underline{$88.24$} & $83.80$ & $82.15$  \\$\lambda_{\mathrm{depth}}=3.5$        & $88.23$ & $\mathbf{83.92}$ & \underline{$82.27$  }\\$\lambda_{\mathrm{depth}}=4.5$      & $88.11$ & $83.88$ & $83.26$  \\
      \bottomrule
    \end{tabular}
  }
  \label{tab:ablation2}
\end{table}

\begin{table}[htbp]
  \caption{Sensitivity analysis of the CPS value.}
  \centering
  \resizebox{0.8\columnwidth}{!}{
    \begin{tabular}{l|ccc}
      \toprule      
      \textbf{Method}   & 8 tasks           & 14 tasks          & 20 tasks \\ 
      \midrule
      % Ours (Adaptive CPS)        & $\mathbf{88.36}$ & $\mathbf{83.98}$ & $\mathbf{82.15}$ \\ \midrule CPS fixed at $0.25$   & $88.08$ & $83.50$ & $81.76$  \\CPS fixed at $0.50$        & $88.19$ & $83.73$ & $81.94$  \\CPS fixed at $0.75$      & $88.25$ & $83.68$ & $82.03$  \\CPS fixed at $1.00$      & $87.94$ & $83.76$ & $82.09$  \\
      Ours (Adaptive CPS)        & $\mathbf{88.36}$ & $\mathbf{83.89}$ & $\mathbf{82.30}$ \\ \midrule CPS fixed at $0.25$   & $88.03$ & $83.52$ & $81.78$  \\CPS fixed at $0.50$        & $88.16$ & $83.73$ & $82.00$  \\CPS fixed at $0.75$      & \underline{$88.17$} & $83.66$ & $82.12$  \\CPS fixed at $1.00$      & $87.92$ & \underline{$83.79$} & \underline{$82.19$}  \\
      \bottomrule
    \end{tabular}
  }
  \label{tab:ablation3}
\end{table}

\noindent \textbf{Sensitivity to CPS.}
Finally, we compare the proposed adaptive CPS with fixed CPS values in Tab.~\ref{tab:ablation3}. Adaptive CPS consistently achieves the best performance across all three task scales. Fixed CPS values cannot simultaneously accommodate tasks with different spectral concentration: small values overly preserve rigid task spectra, whereas large values impose stronger spectral modification regardless of task-specific structure. In contrast, adaptive CPS derives the transformation strength from each task's spectral concentration, allowing spectrally rigid tasks to remain close to their original updates while permitting more distributed tasks to undergo stronger adaptation. The consistent advantage of adaptive CPS therefore directly supports the task-aware spectral adaptation in SADA-Merging.

\section{Conclusion}

We presented SADA-Merging, a data-free model merging framework that explicitly accounts for task- and depth-dependent spectral heterogeneity. It adaptively allocates subspace capacity based on spectral complexity, adjusts retained spectra according to task plasticity, and anchors task contributions according to spectral rigidity and network depth. Experiments under both FFT and LoRA settings show consistent improvements over existing merging methods across architectures and task scales, with particularly clear gains for tasks with heterogeneous spectral plasticity. Ablation studies further confirm the complementary roles of spectral adaptation, elastic rank allocation, and depth-dependent anchoring. Notably, experiments on unseen tasks show that SADA-Merging can improve performance beyond the tasks used for merging, suggesting that adaptive subspace fusion can provide a degree of generalization to previously unseen tasks.

\noindent \textbf{Limitations and Future Works.} The gains of SADA-Merging are more limited in the LoRA setting, where task updates exhibit weaker spectral heterogeneity and offer less room for task-adaptive modeling. Future work could explore merging strategies with low-heterogeneity parameter-efficient updates. More broadly, the observed unseen-task generalization raises an open question of how to construct merged subspaces that provide persistent generalization beyond the observed task set, potentially through more explicit modeling of task interactions and subspace structure.
{
    \small
    \bibliographystyle{ieeenat_fullname}
    \bibliography{main}
}
\clearpage
\setcounter{page}{1}
\maketitlesupplementary

\begin{appendix}

\noindent Our supplementary material is organized into the following sections:
\begin{itemize}
    \item Sec. \ref{secA} presents the implementation details of SADA-Merging, its computational and memory cost compared with existing methods, and a dedicated analysis of the LoRA setting.
    \item Sec. \ref{secB} provides additional experimental details, reports unseen-task generalization results across all evaluated backbones, and presents extended ablation studies and sensitivity analysis.
    \item Sec. \ref{B4} presents visualizations of the proposed method and its main experimental results.
\end{itemize}

\section{Algorithm Details}
\label{secA}

\subsection{Implementation Details}
\label{A1}

\noindent \textbf{Elastic Subspace Allocation.} The rank-allocation mapping $\mathcal{R}(\cdot)$ in Sec.~\ref{sec:elastic} converts standardized spectral entropy into a non-negative rank increment. Specifically, for task $i$ at layer $l$, we define
\begin{equation}
\mathcal{R}(\hat{H}_{i,l})
=
\max\left\{
0,\,
\left\lfloor
s_N \hat{H}_{i,l}
\right\rfloor
\right\},\hat{H}_{i,l}
=
\frac{H_{i,l}-\mu_l}
{\sigma_l+\epsilon}.
\end{equation}
where $H_{i,l}$ denotes the spectral entropy, while $\mu_l$ and $\sigma_l$ are the mean and standard deviation of the task-wise spectral entropies at layer $l$, respectively. The scaling factor $s_N$ controls the magnitude of the entropy-induced rank variation and is set to $12$, $10$, and $8$ for $N=8$, $14$, and $20$, respectively. Larger $s_N$ allows relative differences in spectral complexity to induce larger rank variations, whereas a smaller $s_N$ provides a more conservative allocation when merging a larger number of tasks. The final task-specific rank is obtained by adding this entropy-dependent increment to a shared base rank $r_{\mathrm{base}}=\lfloor d/N\rfloor$, where $d$ is the matrix dimension.

\noindent \textbf{CPS-guided Spectral Adaptation.} For the retained singular values
$\Sigma_{i,l}^{r}
=
\operatorname{diag}(\sigma_{i,l,1},\ldots,\sigma_{i,l,r_{i,l}})$,
we construct the smoothed spectrum $\widetilde{\Sigma}_{i,l}$ by redistributing the retained singular-value mass according to a normalized, monotonically decreasing reference profile $\mathbf{q}_{i,l}$. Specifically, $q_{i,l,j}
=
\frac{r_{i,l}-j+1}
{\sum_{k=1}^{r_{i,l}}(r_{i,l}-k+1)}$. The smoothing strength $\alpha_{i,l}$ is determined by the spectral decay measured at the reference position $k_0=\min(8,d)$, using a linear mapping from the ratio $\sigma_{i,l,1}/\sigma_{i,l,k_0}$ to the smoothing strength. The smoothed profile is then obtained by interpolating between the reference profile and a uniform distribution:
\begin{equation}
\widetilde{\mathbf{q}}_{i,l}
=
(1-\alpha_{i,l})\mathbf{q}_{i,l}
+
\alpha_{i,l}\frac{\mathbf{1}}{r_{i,l}},
\end{equation}
which yields the smoothed singular values
\begin{equation}
\widetilde{\sigma}_{i,l,j}
=
\left(
\sum_{k=1}^{r_{i,l}}\sigma_{i,l,k}
\right)
\widetilde{q}_{i,l,j}.
\end{equation}
The resulting spectrum is subsequently rescaled according to its spectral position and blended with the retained spectrum according to the corresponding CPS score, allowing the degree of spectral adaptation to vary continuously with the rigidity-plasticity characteristics of each task and layer.

% The ECR prefix $K$ is set to $10\%$ of the available singular spectrum, with $\tau=0.5$ and $\beta=20$. The prefix ratio is defined relative to the spectral dimension rather than as an absolute rank, making the concentration measure comparable across layers and architectures.

The ECR prefix $K$ is set to $10\%$ of the available singular spectrum, with $\tau=0.5$ and $\beta=20$. The $10\%$ prefix provides better task-level discrimination in spectral concentration, making the resulting ECR values more effective for distinguishing task-specific rigidity. The prefix ratio is defined relative to the spectral dimension rather than as an absolute rank, making the concentration measure comparable across layers and architectures.

\noindent \textbf{Shared Cover-Space Fusion.} %The retained task subspaces are modulated according to CPS before constructing the shared cover space, using a fixed linear weighting rule. This gives spectrally plastic tasks a slightly larger contribution when defining the shared representation, while applying the same weighting scheme across tasks and layers. The shared cover space is then constructed from the concatenated retained subspaces, so its dimensionality follows directly from the elastic ranks rather than from an independently specified global rank.
The retained task subspaces are first modulated by a fixed CPS-dependent weight and concatenated to form the shared left and right subspaces. Specifically, the weighted left and right subspaces are separately factorized via SVD, and the resulting SVD factors are combined to construct the orthogonal cover matrices. The resulting cover-space coordinates are partitioned according to the cumulative elastic ranks, with $b_i=\sum_{j=1}^i r_{j,l}$, yielding task-associated coordinate blocks $[b_{i-1}:b_i]$. These blocks are then used to extract the coordinates assigned to each task from the shared cover matrices and project the corresponding task subspaces into these coordinates for subsequent block-wise fusion.

\noindent \textbf{Task- and Depth-dependent Anchoring.} The rigidity-aware anchoring coefficient is fixed to $\lambda_{\mathrm{rigid}}=0.40$, providing a moderate preservation bias for spectrally concentrated updates without overriding the CPS-based adaptation. The depth coefficient $\lambda_{\mathrm{depth}}=2.50$ is fixed at the backbone level and reused across all task settings.  The normalized layer depth is $d_l=l/(L-1)$, making the depth-dependent term independent of the absolute number of layers and providing a consistent scale across architectures.

Table~\ref{tab:sada_parameters} summarizes the fixed implementation settings required to reproduce SADA-Merging. 

\begin{table}[t]
\centering
\caption{Fixed implementation settings of SADA-Merging.}
\label{tab:sada_parameters}
\resizebox{\columnwidth}{!}{
\begin{tabular}{lll}
\toprule
\textbf{Parameter} & \textbf{Setting} & \textbf{Role} \\
\midrule
$r_\mathrm{base}$ & $d/N$  & Base rank \\
$s_N$ & $12/10/8$ for $N=8/14/20$ & Entropy-to-rank scaling \\
$K$ & $10\%$ & ECR prefix ratio \\
$\tau,\beta$ & $0.5,\ 20$ & CPS transition \\
$k_0$ & $\min(8,d)$ & Spectral-decay reference \\
$f_{\max}$ & $0.95$ & Maximum spectral fidelity \\
$\lambda_{\mathrm{rigid}}$ & $0.40$ & Rigidity-aware anchoring \\
$\lambda_{\mathrm{depth}}$ & $2.50$ & Depth anchoring\\
\bottomrule
\end{tabular}
}
\end{table}

\subsection{Computational Overhead}
\label{secA:complexity}
\label{A2}

\begin{table*}[htbp]
  \caption{Computational and memory overhead of model merging across different scales and vision backbones. Each metric is reported separately, including runtime (min), peak CPU memory (GB), and peak GPU memory (GB).}
  \centering
  \resizebox{1.0\textwidth}{!}{
  \begin{tabular}{ll|ccc|ccc|ccc}
    \toprule
\multirow{2}{*}{\textbf{Model}} & 
\multirow{2}{*}{\textbf{Method}}
& \multicolumn{3}{c|}{\textbf{8 tasks}}
& \multicolumn{3}{c|}{\textbf{14 tasks}}
& \multicolumn{3}{c}{\textbf{20 tasks}} \\

&
&
\textbf{Time (min)} & \textbf{CPU (GB)} & \textbf{GPU (GB)}
&
\textbf{Time (min)} & \textbf{CPU (GB)} & \textbf{GPU (GB)}
&
\textbf{Time (min)} & \textbf{CPU (GB)} & \textbf{GPU (GB)} \\
    
    \midrule
    
    \multirow{4}{*}{ViT-B-32}
    & TSV-M 
    & 0.31 & 8.09 & 0.93
    & 0.51 & 13.16 & 0.93
    & 0.70 & 18.29 & 0.93 \\
    
    & ISO-CTS
    & 0.43 & 8.16 & 10.11
    & 0.66 & 13.24 & 10.11
    & 0.87 & 18.29 & 10.11 \\
    
    & DC-Merge
    & 0.30 & 8.18 & 1.73
    & 0.49 & 13.26 & 2.32
    & 0.67 & 18.37 & 2.91 \\
    
    \rowcolor{gray!15}
    & Ours
    & 0.31 & 8.21 & 2.38
    & 0.49 & 13.27 & 3.53
    & 0.68 & 18.39 & 4.67 \\
    
    \midrule
    
    \multirow{4}{*}{ViT-B-16}
    & TSV-M 
    & 0.30 & 8.05 & 0.92
    & 0.49 & 13.01 & 0.92
    & 0.67 & 18.05 & 0.92 \\
    
    & ISO-CTS
    & 0.44 & 8.04 & 10.10
    & 0.65 & 12.99 & 10.10
    & 0.87 & 17.99 & 10.10 \\
    
    & DC-Merge
    & 0.31 & 8.07 & 1.72
    & 0.49 & 13.12 & 2.32
    & 0.68 & 18.07 & 2.91 \\
    
    \rowcolor{gray!15}
    & Ours
    & 0.32 & 8.10 & 2.38
    & 0.49 & 13.07 & 3.52
    & 0.67 & 18.10 & 4.67 \\
    
    \midrule
    
    \multirow{4}{*}{ViT-L-14}
    & TSV-M
    & 0.99 & 22.59 & 2.03
    & 1.59 & 37.91 & 2.03
    & 2.13 & 53.27 & 2.03 \\
    
    & ISO-CTS
    & 1.27 & 22.64 & 11.26
    & 1.92 & 38.01 & 11.26
    & 2.66 & 53.31 & 11.26 \\
    
    & DC-Merge
    & 0.96 & 22.74 & 3.25
    & 1.53 & 38.02 & 4.15
    & 2.19 & 53.38 & 5.04 \\
    
    \rowcolor{gray!15}
    & Ours
    & 0.95 & 22.69 & 4.20
    & 1.56 & 38.03 & 5.91
    & 2.20 & 53.43 & 7.65 \\
    
    \bottomrule
  \end{tabular}
  }
  \label{tab:time_memory_overhead}
\end{table*}
SADA-Merging is a data-free post-hoc merging method and introduces additional computation only during the merging stage. Its computational cost mainly arises from three operations: task-wise SVD decomposition, construction of the shared cover space, and projection of task updates into the resulting common coordinate system. For a matrix-valued layer of size $d_{\mathrm{out}}\times d_{\mathrm{in}}$, the initial SVDs constitute the dominant cost when dense SVD is used. Importantly, the subsequent cover-space construction operates only on the retained singular subspaces, whose dimension is determined by the sum of the allocated ranks rather than the full matrix dimension.

Let $d_{\mathrm{out}}$ and $d_{\mathrm{in}}$ denote the dimensions of a matrix-valued layer, $m_l=\min(d_{\mathrm{out}},d_{\mathrm{in}})$, and $R_l=\sum_i r_{i,l}$ denote the total retained rank across the $N$ tasks. Assuming dense SVD, the decomposition of the $N$ task updates requires
$O(Nd_{\mathrm{out}}d_{\mathrm{in}}m_l)$ operations. Constructing the shared cover space from the retained left and right singular subspaces requires SVDs on matrices with $R_l$ columns, resulting in $O((d_{\mathrm{out}}+d_{\mathrm{in}})R_l^2)$ operations. Projecting the task updates into the shared coordinate system requires $O(Nd_{\mathrm{out}}d_{\mathrm{in}}R_l)$ operations. Thus, the per-layer complexity can be written as
\begin{equation}
O\!\left(
Nd_{\mathrm{out}}d_{\mathrm{in}}m_l
+
Nd_{\mathrm{out}}d_{\mathrm{in}}R_l
+
(d_{\mathrm{out}}+d_{\mathrm{in}})R_l^2
\right).
\end{equation}
For approximately square ViT linear layers with $d_{\mathrm{out}}\approx d_{\mathrm{in}}\approx d$, this becomes
\begin{equation}
O\!\left(
Nd^3+Nd^2R_l+dR_l^2
\right).
\end{equation}
Since the retained rank is substantially smaller than the full layer dimension in our setting, the latter two terms are controlled by the adaptive rank budget. In the worst case, the complexity is therefore $O(Nd^3)$, exhibiting linear scaling with the number of merged tasks and placing SADA-Merging in the same task-linear dense-SVD complexity regime as existing subspace-based merging methods such as TSV-M \cite{tsv}, Iso-CTS \cite{marczak2025notaskleftbehind}, and DC-Merge \cite{zhang2026dc}.

We further evaluate the practical computational overhead of SADA-Merging across different model sizes and numbers of merged tasks, comparing it with TSV-M, Iso-CTS, and DC-Merge. We measure wall-clock merging time, peak CPU memory, and peak GPU memory under the same hardware and software environment: an Intel i9-13900KF CPU (24 physical cores, 32 threads), 64 GB RAM, 96 GB swap, and one RTX 4090 with 24 GB VRAM.

Tab.~\ref{tab:time_memory_overhead} summarizes the resulting runtime and memory consumption. Overall, SADA-Merging incurs a computational cost comparable to existing subspace-based methods across all three backbones and different task scales. Its runtime remains close to that of TSV-M, Iso-CTS, and DC-Merge, indicating that the additional spectral adaptation, elastic rank allocation, and depth-dependent anchoring introduce only limited practical overhead.

The merging time increases approximately with the number of merged tasks, consistent with the task-wise decomposition used by SADA-Merging. For example, on ViT-B-32, SADA-Merging requires $0.31$, $0.49$, and $0.68$ minutes for $8$, $14$, and $20$ tasks, respectively. Similar scaling behavior is observed for ViT-B-16 and ViT-L-14. In comparison with DC-Merge, the runtime differences remain small across the evaluated settings, placing both methods within a similar practical computational regime.

For memory consumption, peak CPU memory is primarily associated with storing the task updates and intermediate representations, and therefore generally increases with the number of merged tasks. SADA-Merging follows the same trend as the compared methods. Its peak GPU memory is somewhat higher than that of TSV-M and DC-Merge, mainly because spectral processing and shared cover-space construction require additional intermediate representations. Nevertheless, the memory requirement remains well within the capacity of a single RTX 4090 in all evaluated settings. Even for $20$ tasks on ViT-L-14, SADA-Merging requires only $7.65$ GB of peak GPU memory, demonstrating that the method remains computationally practical for the considered model and task scales.

\subsection{Analysis of the LoRA Setting}
\label{A3}

\begin{figure}[ht]
     \centering
     \includegraphics[width=0.8\linewidth]{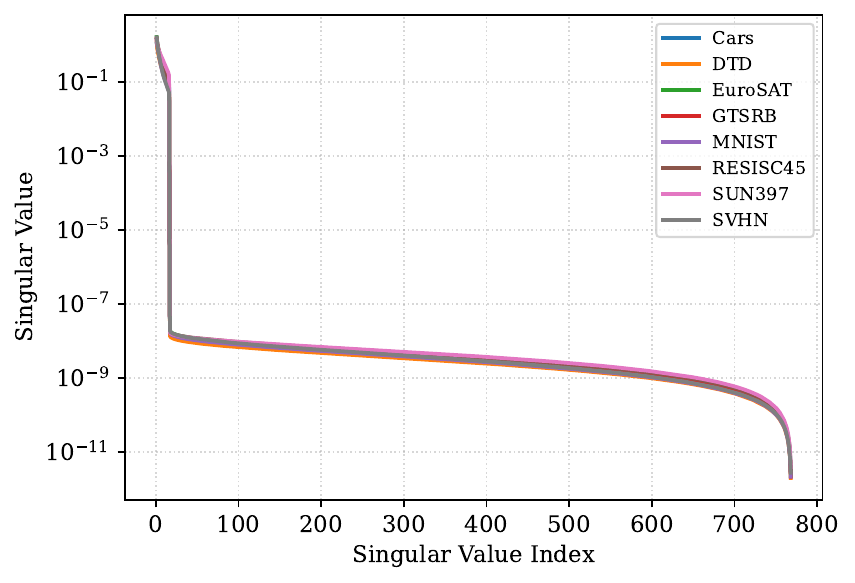}
     \caption{The singular value distributions of task vectors under the LoRA setting.}
     \label{fig:lora1}
\end{figure}

\begin{figure}[ht]
     \centering
     \includegraphics[width=0.8\linewidth]{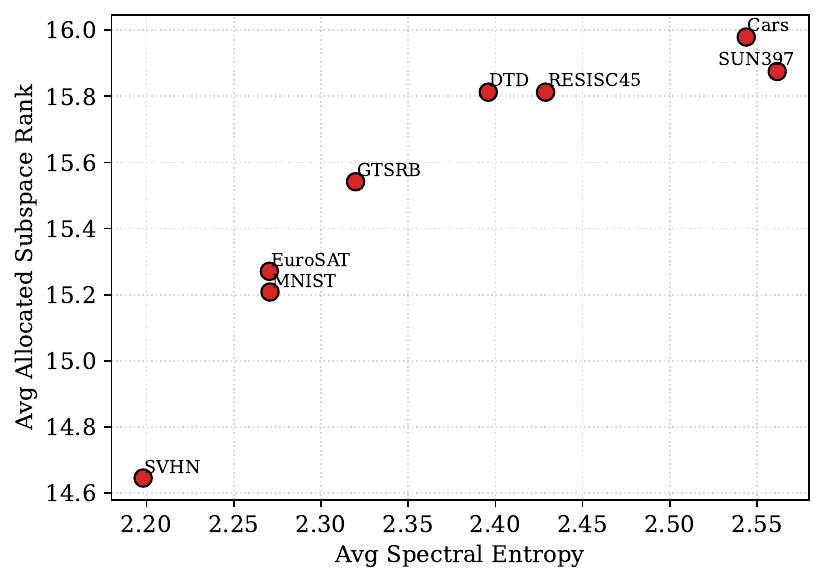}
     \caption{Relationship between spectral entropy and the allocated rank under the LoRA setting.}
     \label{fig:lora2}
\end{figure}

LoRA task vectors exhibit markedly weaker spectral heterogeneity than their FFT counterparts, as shown in Fig.~\ref{fig:lora1}. Nevertheless, spectral entropy still provides sufficient variation to distinguish task-specific rank requirements, with noticeable differences in the allocated ranks across tasks (Fig.~\ref{fig:lora2}). Thus, LoRA updates retain some degree of heterogeneity in terms of spectral rank complexity, allowing the elastic rank allocation mechanism to remain effective.

\begin{figure}[ht]
     \centering
     %\includegraphics[width=0.9\linewidth]{figs/lora_ViT-B-32_8/fig2a_rigidity_plasticity_top0.15.pdf}
     %\caption{Task-wise ECR and CPS under the LoRA setting for $K=15\%$.}
    \includegraphics[width=0.9\linewidth]{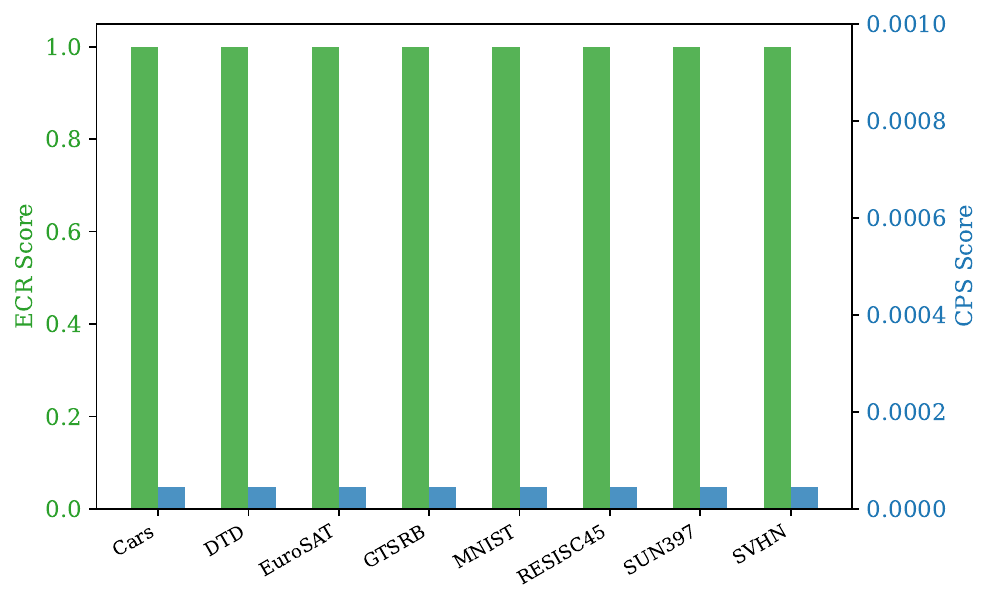}
     \caption{Task-wise ECR and CPS under the LoRA setting for $K=10\%$.}
     \label{fig:lora3}
\end{figure}

\begin{figure}[ht]
     \centering
     % \includegraphics[width=0.9\linewidth]{figs/lora_ViT-B-32_8/fig2a_rigidity_plasticity_top0.01.pdf}
     % \caption{Task-wise ECR and CPS under the LoRA setting for $K=1\%$.}
     \includegraphics[width=0.9\linewidth]{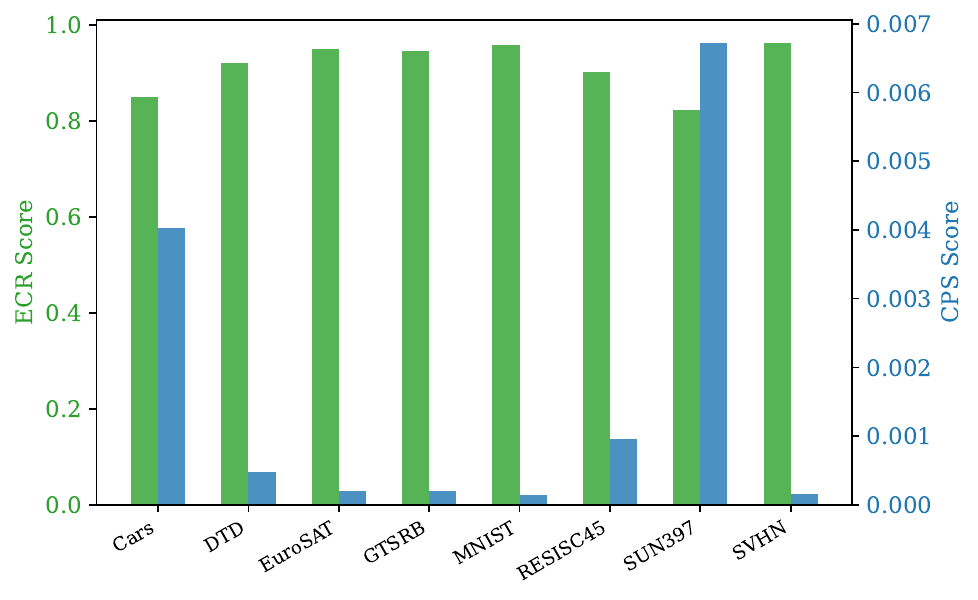}
     \caption{Task-wise ECR and CPS under the LoRA setting for $K=1\%$.}
     \label{fig:lora4}
\end{figure}

In contrast, the ECR-based rigidity-plasticity analysis reveals a much weaker form of task heterogeneity. As shown in Fig.~\ref{fig:lora3}, the ECR values are consistently high across tasks, resulting in CPS values close to zero when $K$ corresponds to the top $10\%$ of singular values. Reducing $K$ to the top $1\%$ does not substantially increase the task-wise variation, with CPS values remaining on the order of $10^{-3}$ (Fig.~\ref{fig:lora4}). This indicates that, although LoRA task vectors can still differ in their effective spectral rank, their spectral concentration is much more homogeneous across tasks. Consequently, the CPS-guided spectral adaptation has limited room to differentiate task-specific updates, which largely explains the more modest gains of SADA-Merging in the LoRA setting. The rank-allocation and depth-dependent anchoring components, however, remain applicable and continue to provide complementary benefits.

\begin{table*}[t]
  \caption{Performance of model merging methods on unseen tasks. Each column follows the \texttt{test:N / train:M} notation, where \texttt{M} of the \texttt{N} tasks are used for merging and the remaining \texttt{N-M} tasks are unseen during merging. The first value indicates performance on unseen tasks, and the second indicates performance on all tasks. \textbf{Bold} indicates the best result, while \underline{underlined} indicates the second-best.}
  \resizebox{1\textwidth}{!}{
    \begin{tabular}{cccc|ccc|ccc}
      \toprule
      \multicolumn{1}{c}{\multirow{2}{*}[-0.5ex]{\textbf{Method}}}         & \multicolumn{3}{c}{ViT-B-32}                     & \multicolumn{3}{c}{ViT-B-16}                     & \multicolumn{3}{c}{ViT-L-14}                    \\ 
                        \cmidrule{2-10}
    & test:14 / train:8  & test:20 / train:8  & test:20 / train:14 
    & test:14 / train:8  & test:20 / train:8  & test:20 / train:14
    & test:14 / train:8  & test:20 / train:8  & test:20 / train:14
    \\ 
                                               \midrule
      \multicolumn{1}{c}{Zeroshot}            & $69.12/57.12$ & $64.08/57.70$ & $59.03/57.70$ & $69.16/61.36$ & $66.49/62.10$ & $63.82/62.10$ & $72.87/68.53$ & $69.70/67.93$ & $66.53/67.93$ \\
      \multicolumn{1}{c}{Individual}            & $88.28/90.87$ & $90.39/91.36$ & $92.51/91.36$ & $90.24/92.75$ & $92.19/93.17$ & $94.13/93.17$ & $92.25/94.28$ & $94.00/94.72$ & $95.74/94.72$ \\ \midrule
      \multicolumn{1}{c}{TSV-M} & $64.22/75.96$ & $\underline{65.48}/72.72$ & $64.38/75.27$ & $66.88/78.75$ & $\underline{67.84}/73.06$ & $66.39/78.88$ & $73.01/84.31$ & $\underline{72.42}/80.29$ & $70.69/83.54$ \\
      
      \multicolumn{1}{c}{Iso-CTS}        & $63.58/76.21$ & $65.02/72.76$ & $64.03/76.23$ & $\underline{67.56}/\underline{80.33}$ & $67.08/76.21$ & $65.90/80.00$ & $71.92/84.94$ & $71.75/80.83$ & $\mathbf{71.60}/\underline{85.19}$ \\
      
      \multicolumn{1}{c}{DC-Merge} & $\underline{64.73}/\underline{77.16}$ & $65.12/\underline{73.48}$ & $\underline{64.78}/\underline{77.16}$ & $67.53/80.13$ & $67.66/\underline{76.43}$ & $\mathbf{67.07}/\underline{80.89}$ & $\underline{72.92}/\underline{85.00}$ & $72.23/\underline{80.91}$ & $71.01/85.11$ \\
      \bottomrule
      
      \rowcolor{gray!15}
      \multicolumn{1}{c}{\textbf{Ours}} &$\mathbf{66.74/78.03}$ & $\mathbf{66.29/74.38}$ & $\mathbf{65.21/78.33}$ & $\mathbf{68.45/80.51}$ & $\mathbf{68.17/76.72}$ & $\underline{66.58}/\mathbf{81.17}$ & $\mathbf{73.26/85.37}$ & $\mathbf{72.51/81.17}$ & $\underline{71.06}/\mathbf{85.30}$ \\
      \bottomrule
    \end{tabular}
  }
  \label{tab:supp_unseen}
\end{table*}

\section{Experimental Details and Additional Results}
\label{secB}

\subsection{Experimental Details}
\label{B1}

For the FFT setting, we consider three benchmark scales. The 8-task benchmark consists of Cars~\cite{cars}, DTD~\cite{dtd}, EuroSAT~\cite{eurosat}, GTSRB~\cite{gtsrb}, MNIST~\cite{MNIST}, RESISC45~\cite{cheng2017remote}, SUN397~\cite{sun397}, and SVHN~\cite{svhn}. The 14-task benchmark retains these eight datasets and includes CIFAR100~\cite{krizhevsky2009learning}, STL10~\cite{coates_analysis_2011}, Flowers102~\cite{nilsback_automated_2008}, OxfordIIITPet~\cite{parkhi_cats_2012}, PCAM~\cite{veeling_rotation_2018}, and FER2013~\cite{goodfellow_challenges_2013}. The 20-task benchmark further incorporates EMNIST~\cite{cohen_emnist_2017}, CIFAR10~\cite{krizhevsky2009learning}, Food101~\cite{bossard_food-101_2014}, FashionMNIST~\cite{xiao_fashion-mnist_2017}, RenderedSST2~\cite{socher_recursive_nodate}, and KMNIST~\cite{clanuwat_deep_2018}.

For the LoRA setting, we use three benchmark scales with 8, 12, and 16 tasks. The 8-task benchmark uses the same dataset set as the FFT setting. Starting from this configuration, the 12-task benchmark includes CIFAR100~\cite{krizhevsky2009learning}, Flowers102~\cite{nilsback_automated_2008}, OxfordIIITPet~\cite{parkhi_cats_2012}, and STL10~\cite{coates_analysis_2011}. The 16-task benchmark additionally incorporates FER2013~\cite{goodfellow_challenges_2013}, CIFAR10~\cite{krizhevsky2009learning}, FashionMNIST~\cite{xiao_fashion-mnist_2017}, and RenderedSST2~\cite{socher_recursive_nodate}.

We additionally evaluated ACE-Merging~\cite{xu2026ace}, a recent data-free model merging approach based on adaptive covariance estimation as a baseline. We implemented ACE-Merging according to the formulation described in the original paper and applied it under the same experimental protocol, including the same task vectors, vision backbones, and evaluation settings used for other merging baselines. However, in our settings, ACE-Merging did not consistently outperform existing competitive baselines across different task scales and architectures. Since the main comparison focuses on representative methods with stable performance across diverse settings, we report the consistently competitive baselines in the main tables and leave ACE-Merging out of the primary comparison. This observation also suggests that covariance-based aggregation alone may not fully characterize the heterogeneous spectral structures of modern vision transformer task updates, motivating the spectrum-aware and adaptive subspace design of SADA-Merging.

\subsection{Unseen Task Generalization}
\label{B2}
We provide additional unseen-task generalization results in Tab.~\ref{tab:supp_unseen}, extending the analysis in the main paper to ViT-B-16 and ViT-L-14, while also reporting the corresponding ViT-B-32 results for completeness. Across the three backbones, we consider different unseen-task configurations to assess the robustness of the observed generalization advantage.

On ViT-B-16, SADA-Merging outperforms DC-Merge in unseen-task accuracy in two of the three settings, with gains of $0.92$ and $0.51$ points when merging 8 tasks for the 14- and 20-task benchmarks, respectively. When merging 14 of 20 tasks, it trails DC-Merge by $0.49$ points on unseen tasks, but still achieves $0.28$ points higher all-task accuracy. ViT-L-14 shows a similar trend. SADA-Merging achieves the highest unseen-task accuracy in the first two settings%, surpassing DC-Merge by $0.34$ and $0.28$ points, respectively, 
and attains the best all-task performance in all three settings.

Overall, SADA-Merging achieves the best all-task performance in all nine settings and remains among the strongest data-free merging methods across all three backbones. These results indicate that its adaptive subspace allocation and task-preserving fusion generalize beyond ViT-B-32 and remain effective when target tasks are excluded from the merging process.

\subsection{Additional Ablation Study and Sensitivity Analysis}
\label{B3}
\begin{table*}[htbp]
  \caption{Ablation study of SADA-Merging with progressively added components.}
  \centering
  \resizebox{0.9\textwidth}{!}{
    \begin{tabular}{lccc|ccc|ccc}
      \toprule
      \multicolumn{1}{c}{\multirow{2}{*}[-0.5ex]{\textbf{Method}}} & \multicolumn{3}{c}{ViT-B-32} & \multicolumn{3}{c}{ViT-B-16} & \multicolumn{3}{c}{ViT-L-14} \\
      \cmidrule{2-10}
      & 8 tasks & 14 tasks & 20 tasks & 8 tasks & 14 tasks & 20 tasks & 8 tasks & 14 tasks & 20 tasks \\
      \midrule
      % Base & $87.03$ & $82.49$ & $80.38$ & $90.38$ & $86.60$ & $84.05$ & $93.98$ & $90.77$ & $89.98$ \\\midrule+ CPS & $87.42$ & $83.05$ & $80.86$ & $90.68$ & $86.72$ & $84.23$ & $94.25$ & $90.93$ & $90.06$ \\+ CPS + Rank & $87.53$ & $83.21$ & $81.01$ & $90.81$ & $86.95$ & $84.41$ & $94.37$ & $91.11$ & $90.14$ \\+ CPS + Rank + Depth (Ours) & $\mathbf{88.36}$ & $\mathbf{83.98}$ & $\mathbf{82.15}$ & $\mathbf{91.08}$ & $\mathbf{87.41}$ & $\mathbf{85.05}$ & $\mathbf{94.52}$ & $\mathbf{91.37}$ & $\mathbf{90.73}$ \\
      Base & $87.10$ & $82.56$ & $80.54$ & $90.44$ & $86.54$ & $83.95$ & $93.95$ & $90.72$ & $89.96$ \\\midrule+ CPS & $87.37$ & $83.00$ & $80.94$ & $90.74$ & $86.75$ & $84.34$ & $94.27$ & $90.98$ & $90.18$ \\+ CPS + Rank & \underline{$87.53$} & \underline{$83.23$} & \underline{$81.15$} & \underline{ $90.96$} & \underline{$86.94$} & \underline{$84.45$} & \underline{$94.33$} & \underline{$91.06$} & \underline{$90.27$} \\+ CPS + Rank + Depth (Ours) & $\mathbf{88.36}$ & $\mathbf{83.89}$ & $\mathbf{82.30}$ & $\mathbf{91.11}$ & $\mathbf{87.41}$ & $\mathbf{85.10}$ & $\mathbf{94.54}$ & $\mathbf{91.46}$ & $\mathbf{90.85}$ \\
      \bottomrule
    \end{tabular}
  }
  \label{tab:ablationsupp}
\end{table*}

\noindent \textbf{Extended Component Ablation. }We further extend the component-wise ablation study in the main paper to ViT-B-16 and ViT-L-14. The corresponding ViT-B-32 results are included for completeness, while the additional experiments examine whether the contributions of CPS-guided spectral adaptation, elastic rank allocation, and depth-dependent anchoring remain consistent across different backbone architectures.

As shown in Tab.~\ref{tab:ablationsupp}, the progressive addition of the three components yields consistent improvements on both ViT-B-16 and ViT-L-14. On ViT-B/16, introducing CPS improves performance over the Base configuration across all three task scales, and elastic rank allocation provides further gains. Adding depth-dependent anchoring produces the largest improvement.%, reaching $91.11$, $87.41$, and $85.10$ on the 8-, 14-, and 20-task settings, respectively. 
A similar trend is observed on ViT-L/14, consistently outperforming the progressively simplified variants.

These results reinforce the component-level findings in the main paper and show that the benefits of spectrum-aware adaptation, elastic subspace allocation, and depth-dependent anchoring are not specific to ViT-B/32, but generalize across different vision backbones.

\noindent \textbf{Sensitivity to ECR Ratio $K$.} We evaluate the sensitivity of the ECR prefix ratio $K$ used in CPS computation in Tab.~\ref{tab:ablationk}. SADA-Merging exhibits stable performance across a relatively broad range of $K$ values from $0.05$ to $0.20$. On the 8-task benchmark, the absolute accuracy varies by less than $0.25$ points, with both $K=0.10$ and $K=0.15$ achieving $88.36$. Similar stability is observed on the 14- and 20-task benchmarks, where the accuracy varies within $0.23$ and $0.39$ points, respectively. Notably, the default $K=0.10$ achieves the best performance on the 20-task benchmark and remains competitive at the other task scales. Overall, these results indicate that CPS-based spectral adaptation is robust to the precise choice of ECR prefix and does not rely on narrow hyperparameter tuning.

\begin{table}[htbp]
  \caption{Sensitivity analysis of the ECR prefix ratio $K$ used for CPS computation.}
  \centering
  \resizebox{0.9\columnwidth}{!}{
    \begin{tabular}{l|ccc}
      \toprule
      \textbf{Method} & 8 tasks & 14 tasks & 20 tasks \\
      \midrule
      Ours ($K=0.10$) & $\mathbf{88.36}$ & \underline{$83.89$} & $\mathbf{82.30}$ \\ \midrule
      $K=0.05$ & $88.12$ & $83.84$ & \underline{$82.26$} \\
      $K=0.15$ & $\mathbf{88.36}$ & $\mathbf{83.98}$ & $82.15$ \\
      $K=0.20$ & \underline{$88.28$} & $83.75$ & $81.91$ \\
      \bottomrule
    \end{tabular}
  }
  \label{tab:ablationk}
\end{table}

\noindent \textbf{Sensitivity to Rank Budget.} We evaluate the sensitivity to the layer-wise rank budget in Tab.~\ref{tab:ablationrank}. In addition to the original setting without an explicit budget, we evaluate rank scaling factors of $0.75$, $1.00$, $1.25$, and $1.50$. For each layer, the factor is multiplied by the layer dimension to determine the rank budget, and the task-specific ranks are proportionally adjusted to match the resulting budget.

% Compared with the unconstrained setting, the $0.75$ budget leads to a decrease of at most $0.25$ points. Increasing the budget from $0.75$ to $1.00$ recovers most of this loss, while further increasing it to $1.25$ and $1.50$ yields performance comparable to the unconstrained setting. These results also support the reasonableness of the default no-budget setting: imposing a tighter rank constraint can slightly reduce performance, whereas increasing the capacity beyond the original elastic allocation provides no consistent benefit. Thus, allowing elastic rank allocation to determine the layer-wise rank without an additional hard budget avoids unnecessary capacity constraints while maintaining stable performance. Overall, the results suggest that the effectiveness of SADA-Merging is primarily driven by adaptive rank allocation rather than increased representational capacity.

Compared with the unconstrained setting, the $0.75$ budget leads to a decrease of at most $0.25$ points, while increasing the budget to $1.00$ largely recovers this loss. Further increasing the budget does not yield consistent improvements, with performance beginning to markedly decline at $1.50$. These results support the reasonableness of the default no-budget setting: tighter rank constraints can slightly reduce performance, while excessive rank budgets may introduce unnecessary capacity without improving the quality of the merged subspaces. Thus, allowing elastic rank allocation to determine the layer-wise rank without an additional hard budget avoids unnecessary capacity constraints while maintaining stable performance. Overall, the results suggest that the effectiveness of SADA-Merging is primarily driven by adaptive rank allocation rather than simply increasing representational capacity.

\begin{table}[htbp]
\caption{Sensitivity analysis of the layer-wise rank budget. }
\centering
\resizebox{0.9\columnwidth}{!}{
\begin{tabular}{l|ccc}
\toprule
\textbf{Method} & 8 tasks & 14 tasks & 20 tasks \\
\midrule
Ours (no budget) & $\mathbf{88.36}$ & \underline{$83.89$} & $\mathbf{82.30}$ \\ \midrule
$\text{budget} = 0.75$ & $88.15$ & $83.83$ & $82.05$ \\
$\text{budget} = 1.00$ & \underline{$88.28$} & $\mathbf{84.02}$ & \underline{$82.21$} \\
$\text{budget} = 1.25$ & $88.21$ & \underline{$83.89$} & $82.15$ \\
$\text{budget} = 1.50$ & $88.01$ & $83.79$ & $81.88$ \\
\bottomrule
\end{tabular}
}
\label{tab:ablationrank}
\end{table}

\section{Additional Visualizations}
\label{B4}
\begin{figure*}[htbp]
  \centering
  \begin{subfigure}[b]{0.32\linewidth}
    \centering
    \includegraphics[width=\linewidth]{figs/ViT-B-32_8/fig1a_singular_spectrum.pdf}
    \caption{ViT-B-32, 8 tasks}
    \label{fig:sv_a1}
  \end{subfigure}
  \hfill
  \begin{subfigure}[b]{0.32\linewidth}
    \centering
    \includegraphics[width=\linewidth]{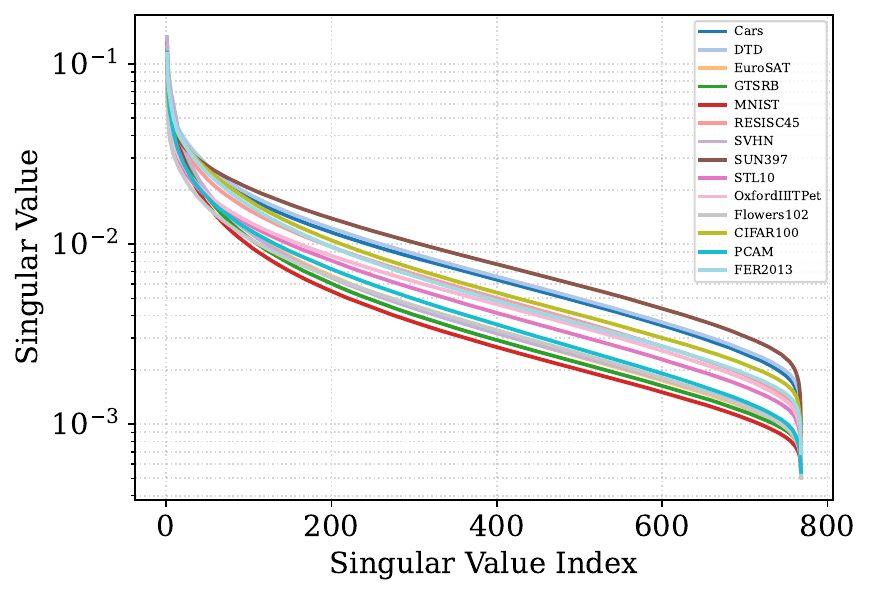}
    \caption{ViT-B-32, 14 tasks}
    \label{fig:sv_a2}
  \end{subfigure}
  \hfill
  \begin{subfigure}[b]{0.32\linewidth}
    \centering
    \includegraphics[width=\linewidth]{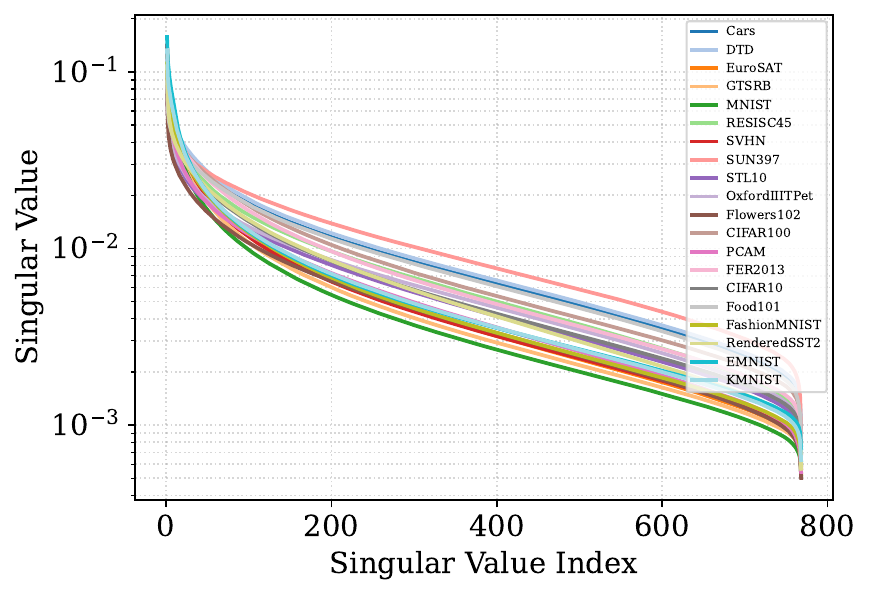}
    \caption{ViT-B-32, 20 tasks}
    \label{fig:sv_a3}
  \end{subfigure}

  \vspace{0.5cm}

  \begin{subfigure}[b]{0.32\linewidth}
    \centering
    \includegraphics[width=\linewidth]{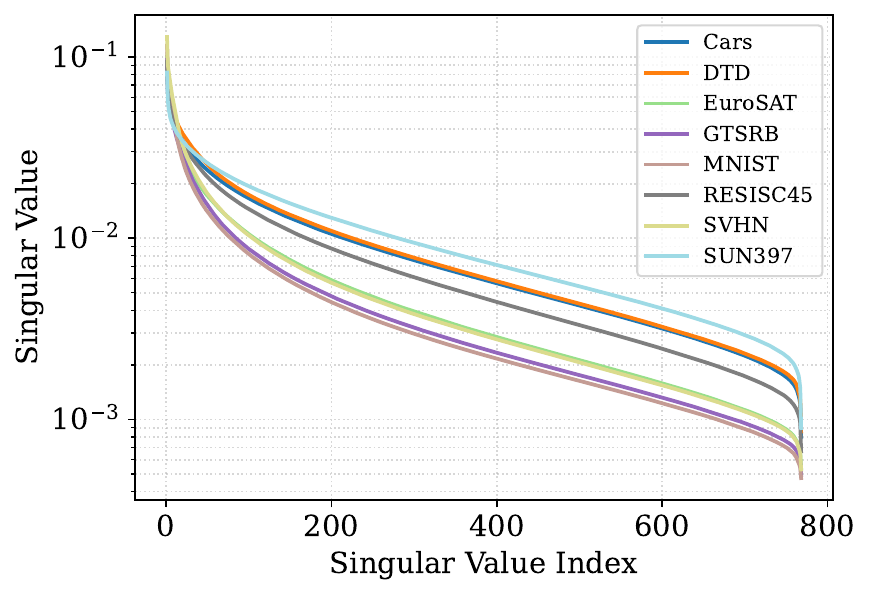}
    \caption{ViT-B-16, 8 tasks}
    \label{fig:sv_b1}
  \end{subfigure}
  \hfill
  \begin{subfigure}[b]{0.32\linewidth}
    \centering
    \includegraphics[width=\linewidth]{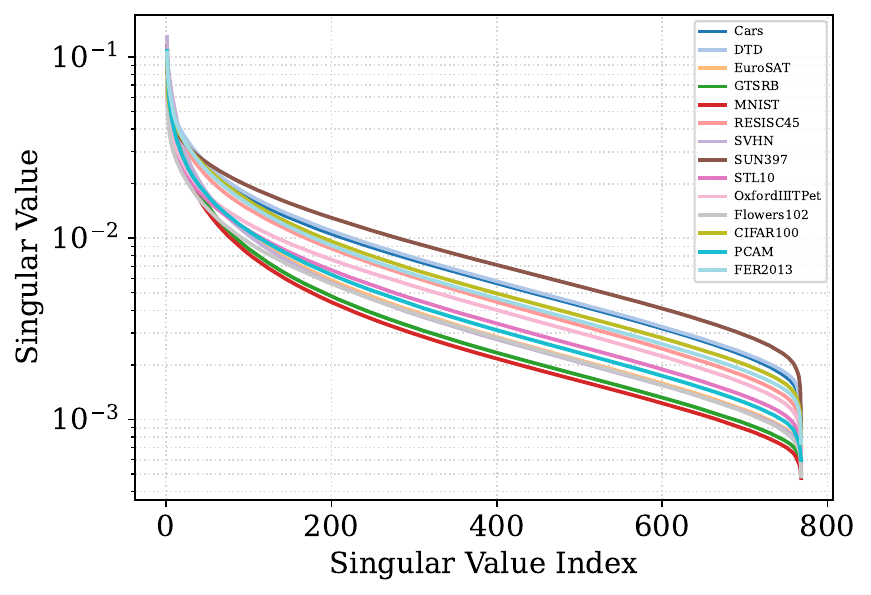}
    \caption{ViT-B-16, 14 tasks}
    \label{fig:sv_b2}
  \end{subfigure}
  \hfill
  \begin{subfigure}[b]{0.32\linewidth}
    \centering
    \includegraphics[width=\linewidth]{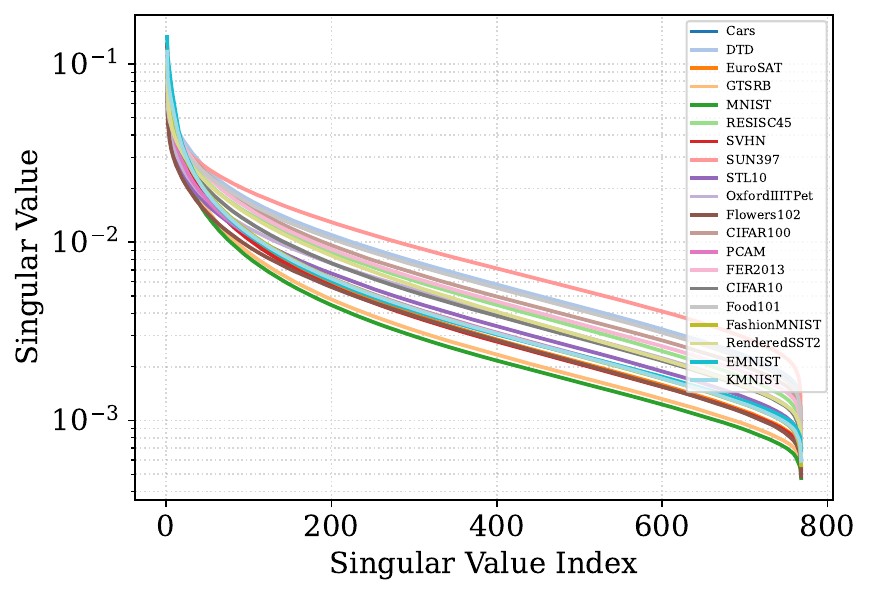}
    \caption{ViT-B-16, 20 tasks}
    \label{fig:sv_b3}
  \end{subfigure}

  \vspace{0.5cm}

  \begin{subfigure}[b]{0.32\linewidth}
    \centering
    \includegraphics[width=\linewidth]{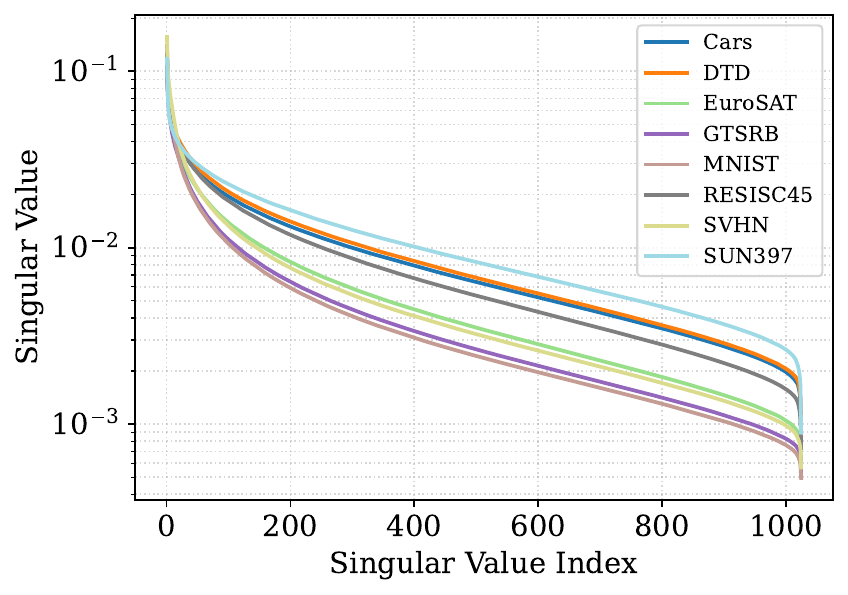}
    \caption{ViT-L-14, 8 tasks}
    \label{fig:sv_c1}
  \end{subfigure}
  \hfill
  \begin{subfigure}[b]{0.32\linewidth}
    \centering
    \includegraphics[width=\linewidth]{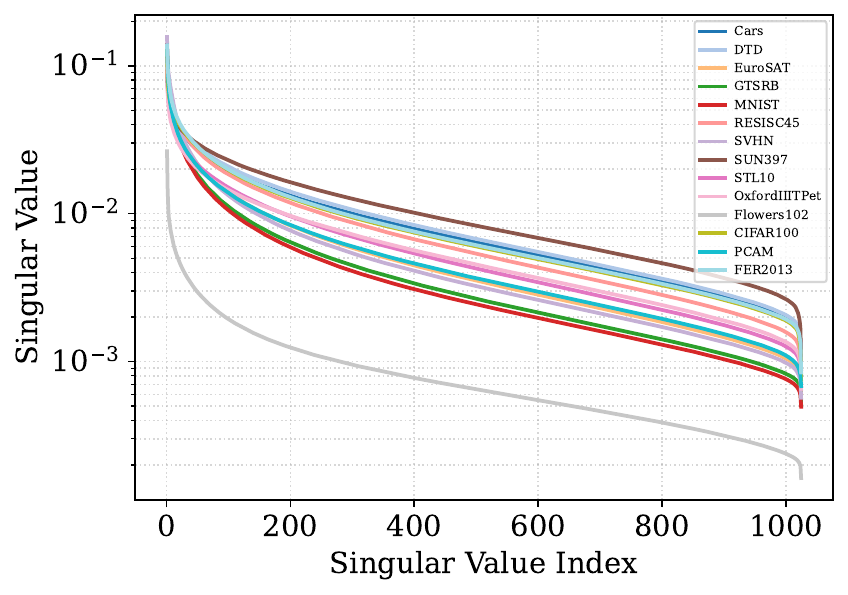}
    \caption{ViT-L-14, 14 tasks}
    \label{fig:sv_c2}
  \end{subfigure}
  \hfill
  \begin{subfigure}[b]{0.32\linewidth}
    \centering
    \includegraphics[width=\linewidth]{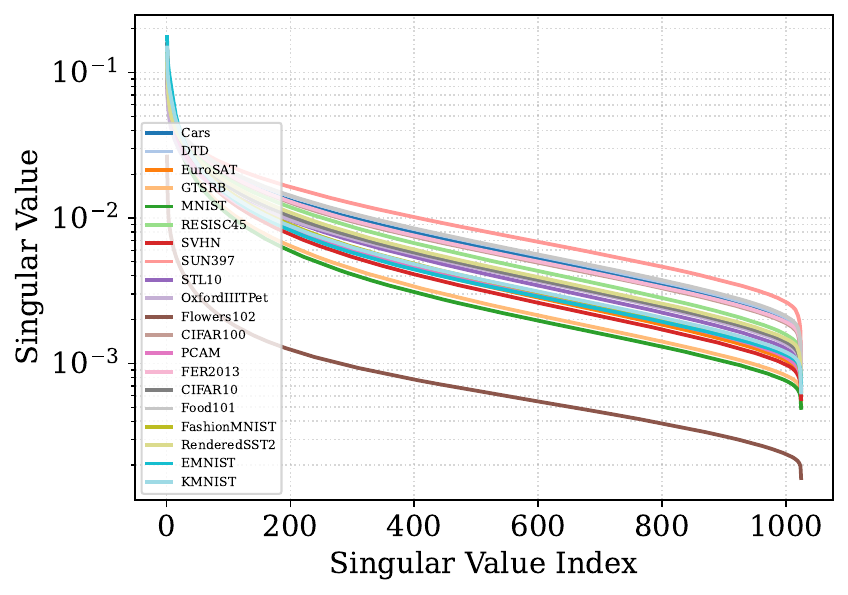}
    \caption{ViT-L-14, 20 tasks}
    \label{fig:sv_c3}
  \end{subfigure}

  \caption{Singular‑value distributions of different vision backbones under the FFT setting.}
  \label{fig:sv_all_backbone}
\end{figure*}

\begin{figure*}[htbp]
  \centering
  \begin{subfigure}[b]{0.32\linewidth}
    \centering
    \includegraphics[width=\linewidth]{figs/ViT-B-32_8/fig1b_entropy_vs_rank.pdf}
    \caption{ViT-B-32, 8 tasks}
    \label{fig:ent_a1}
  \end{subfigure}
  \hfill
  \begin{subfigure}[b]{0.32\linewidth}
    \centering
    \includegraphics[width=\linewidth]{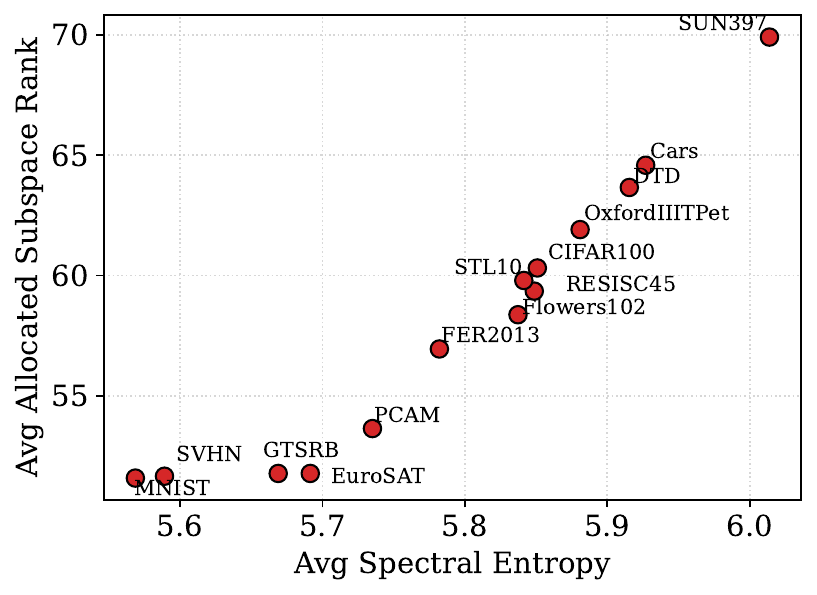}
    \caption{ViT-B-32, 14 tasks}
    \label{fig:ent_a2}
  \end{subfigure}
  \hfill
  \begin{subfigure}[b]{0.32\linewidth}
    \centering
    \includegraphics[width=\linewidth]{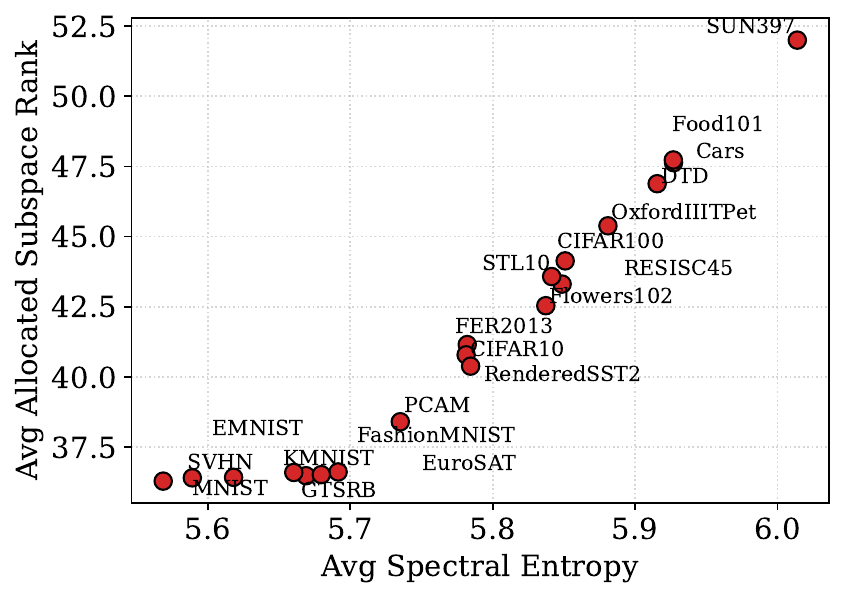}
    \caption{ViT-B-32, 20 tasks}
    \label{fig:ent_a3}
  \end{subfigure}

  \vspace{0.5cm}

  \begin{subfigure}[b]{0.32\linewidth}
    \centering
    \includegraphics[width=\linewidth]{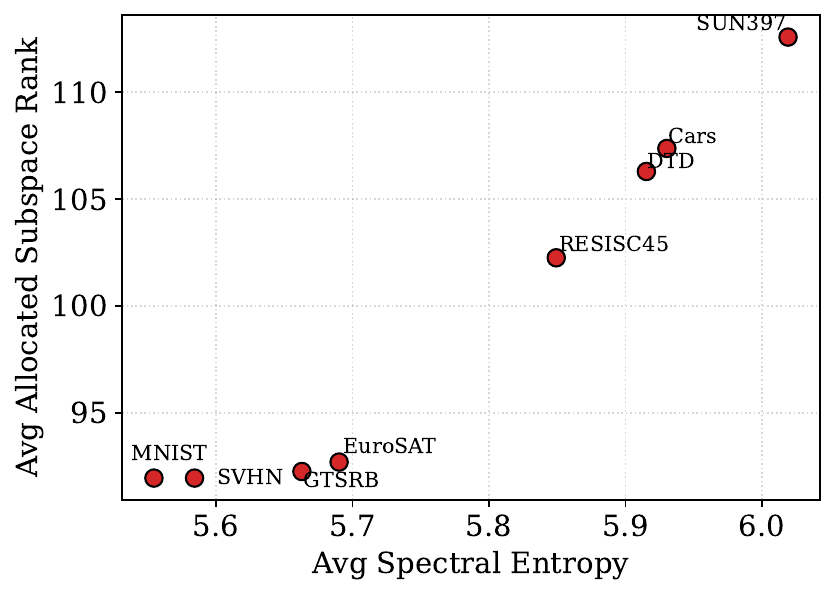}
    \caption{ViT-B-16, 8 tasks}
    \label{fig:ent_b1}
  \end{subfigure}
  \hfill
  \begin{subfigure}[b]{0.32\linewidth}
    \centering
    \includegraphics[width=\linewidth]{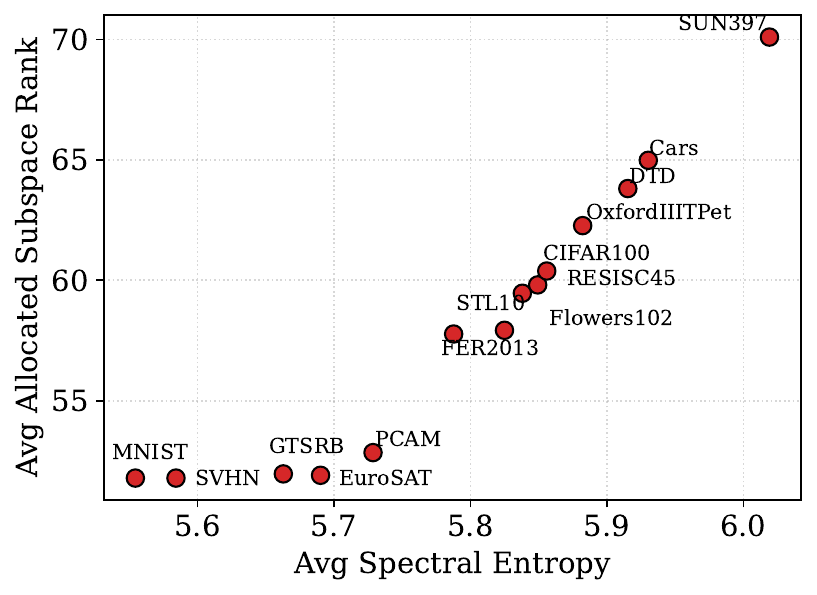}
    \caption{ViT-B-16, 14 tasks}
    \label{fig:ent_b2}
  \end{subfigure}
  \hfill
  \begin{subfigure}[b]{0.32\linewidth}
    \centering
    \includegraphics[width=\linewidth]{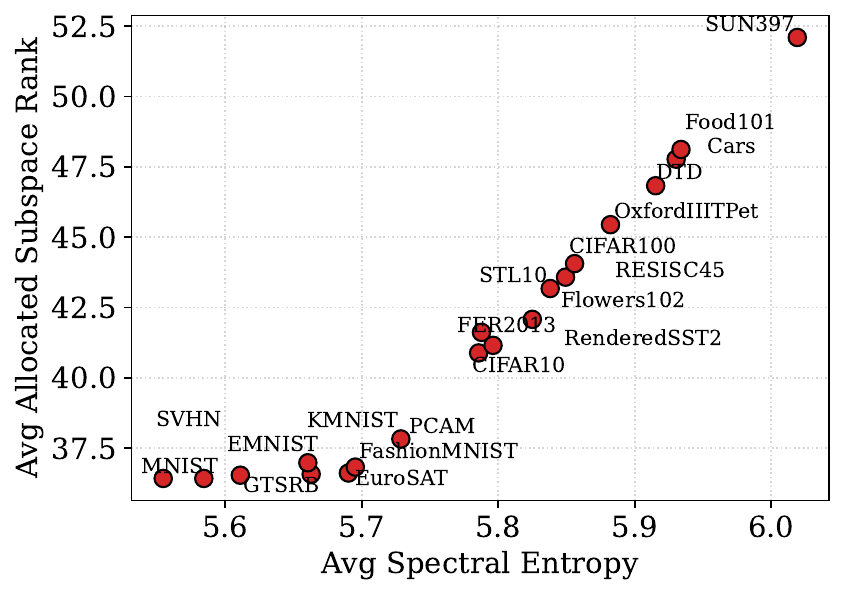}
    \caption{ViT-B-16, 20 tasks}
    \label{fig:ent_b3}
  \end{subfigure}

  \vspace{0.5cm}

  \begin{subfigure}[b]{0.32\linewidth}
    \centering
    \includegraphics[width=\linewidth]{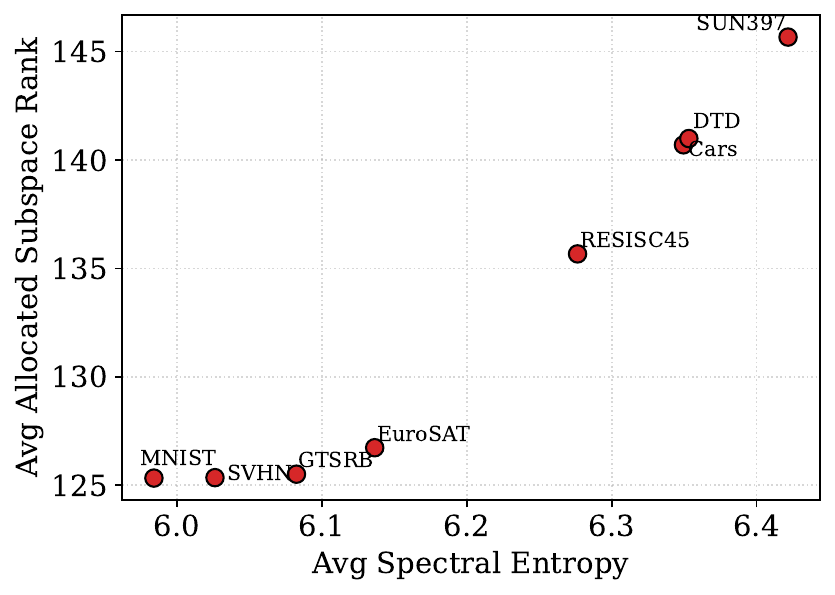}
    \caption{ViT-L-14, 8 tasks}
    \label{fig:ent_c1}
  \end{subfigure}
  \hfill
  \begin{subfigure}[b]{0.32\linewidth}
    \centering
    \includegraphics[width=\linewidth]{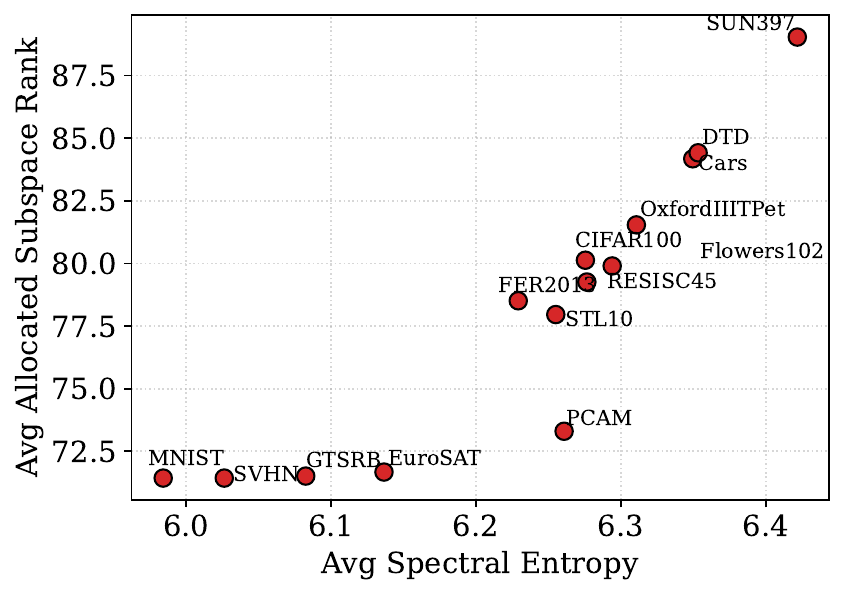}
    \caption{ViT-L-14, 14 tasks}
    \label{fig:ent_c2}
  \end{subfigure}
  \hfill
  \begin{subfigure}[b]{0.32\linewidth}
    \centering
    \includegraphics[width=\linewidth]{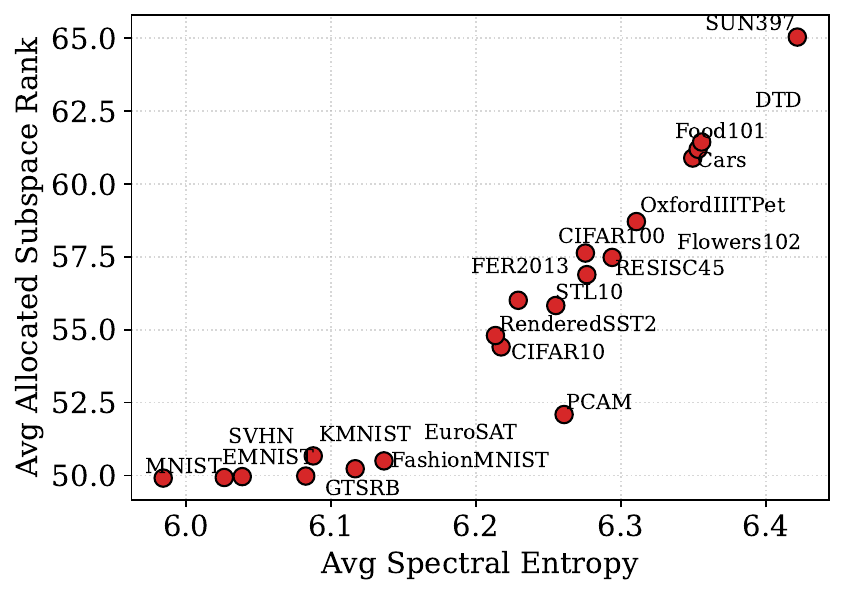}
    \caption{ViT-L-14, 20 tasks}
    \label{fig:ent_c3}
  \end{subfigure}

  \caption{Spectral entropy‑based rank allocation for different vision backbones.}
  \label{fig:entropy_rank_all}
\end{figure*}

\begin{figure*}[htbp]
  \centering
  \begin{subfigure}[b]{0.32\linewidth}
    \centering
    \includegraphics[width=\linewidth]{figs/paper_figs_new/ViT-B-32_8/fig2a_rigidity_plasticity.pdf}
    \caption{ViT-B-32, 8 tasks}
    \label{fig:rp_a1}
  \end{subfigure}
  \hfill
  \begin{subfigure}[b]{0.32\linewidth}
    \centering
    \includegraphics[width=\linewidth]{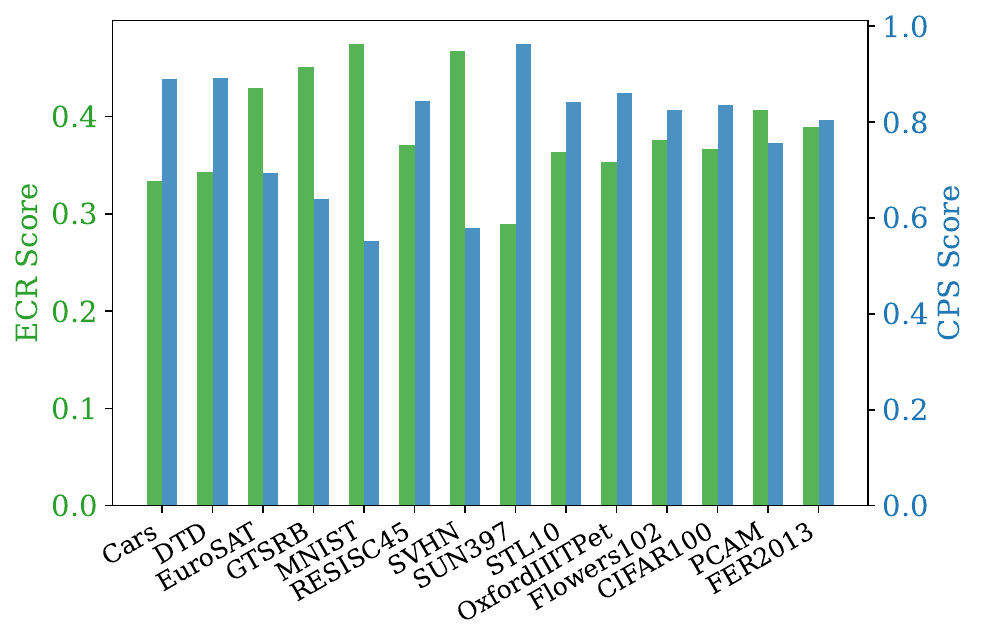}
    \caption{ViT-B-32, 14 tasks}
    \label{fig:rp_a2}
  \end{subfigure}
  \hfill
  \begin{subfigure}[b]{0.32\linewidth}
    \centering
    \includegraphics[width=\linewidth]{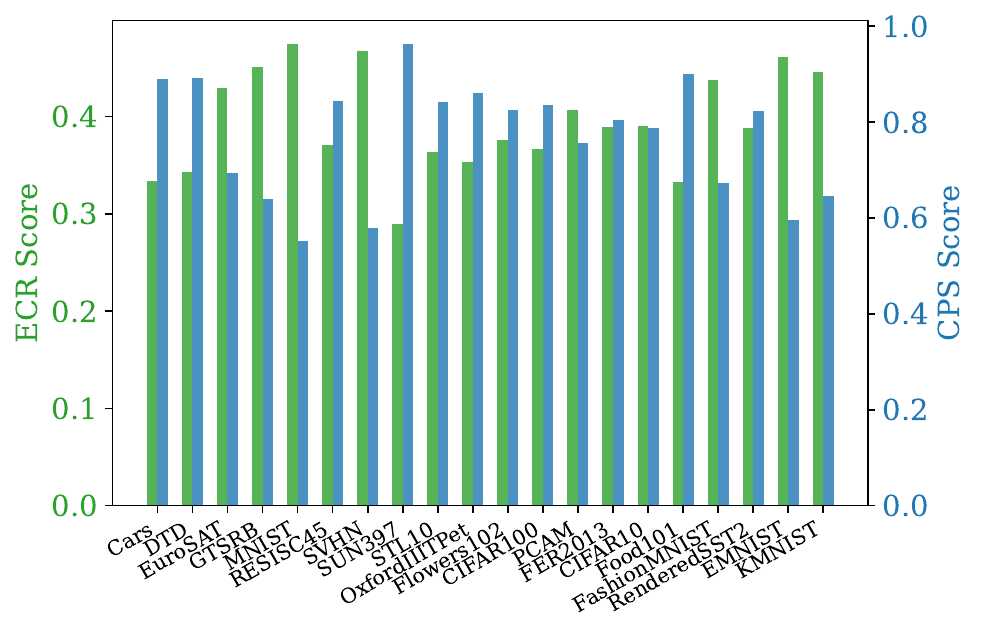}
    \caption{ViT-B-32, 20 tasks}
    \label{fig:rp_a3}
  \end{subfigure}

  \vspace{0.5cm}

  \begin{subfigure}[b]{0.32\linewidth}
    \centering
    \includegraphics[width=\linewidth]{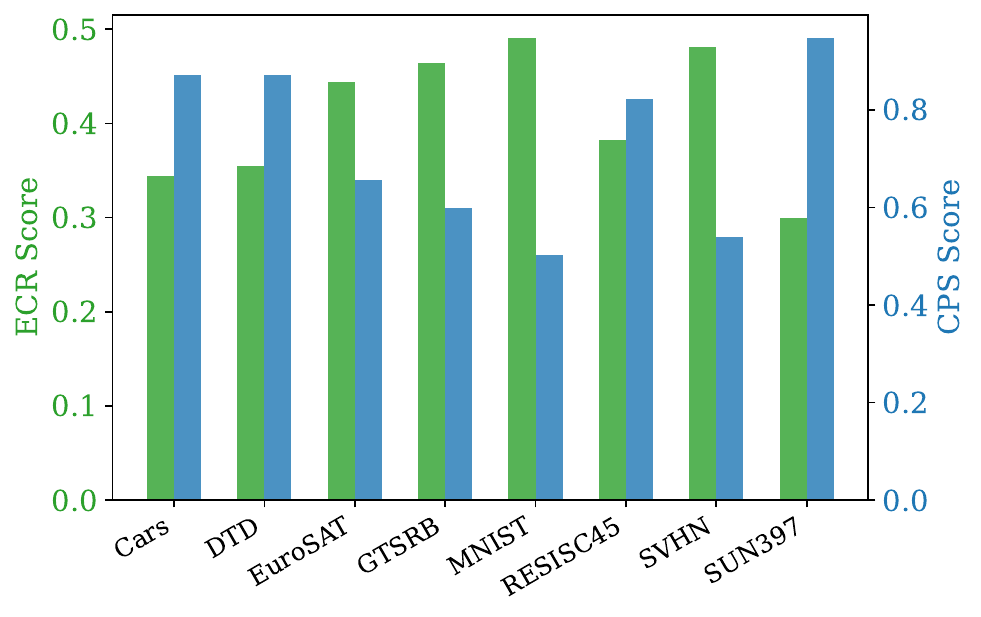}
    \caption{ViT-B-16, 8 tasks}
    \label{fig:rp_b1}
  \end{subfigure}
  \hfill
  \begin{subfigure}[b]{0.32\linewidth}
    \centering
    \includegraphics[width=\linewidth]{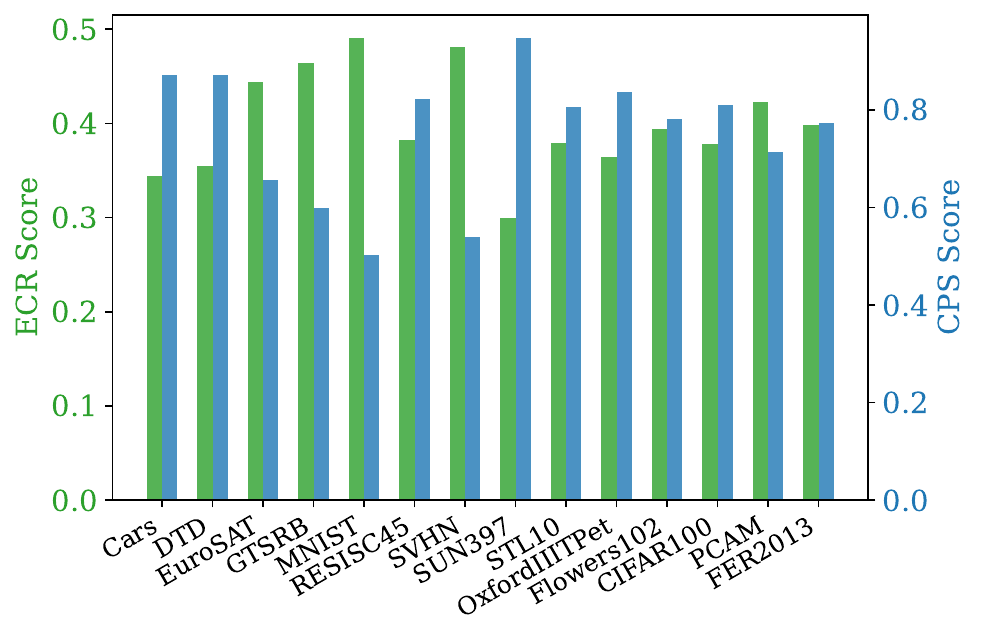}
    \caption{ViT-B-16, 14 tasks}
    \label{fig:rp_b2}
  \end{subfigure}
  \hfill
  \begin{subfigure}[b]{0.32\linewidth}
    \centering
    \includegraphics[width=\linewidth]{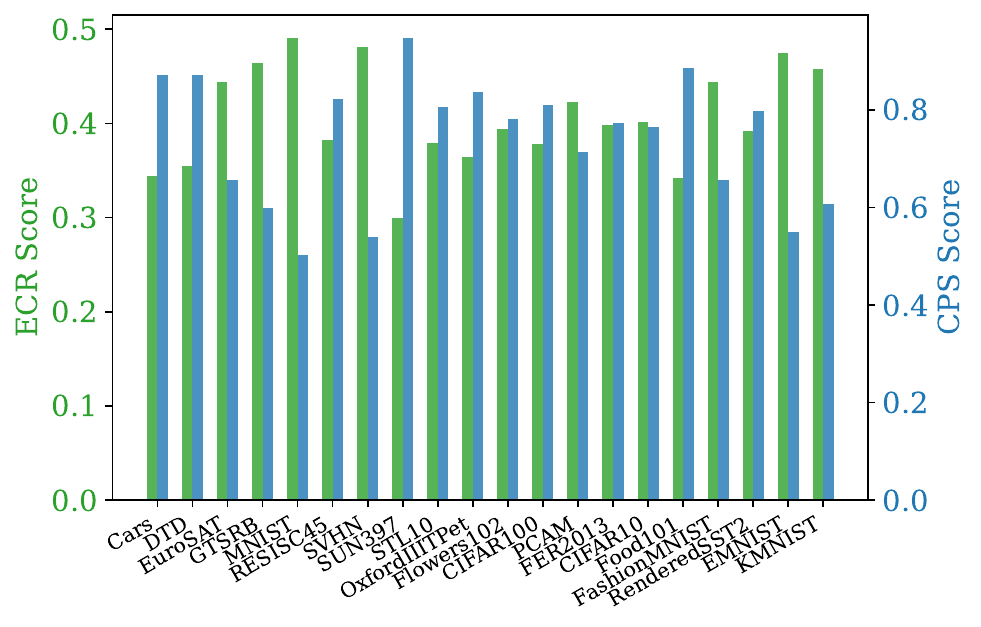}
    \caption{ViT-B-16, 20 tasks}
    \label{fig:rp_b3}
  \end{subfigure}

  \vspace{0.5cm}

  \begin{subfigure}[b]{0.32\linewidth}
    \centering
    \includegraphics[width=\linewidth]{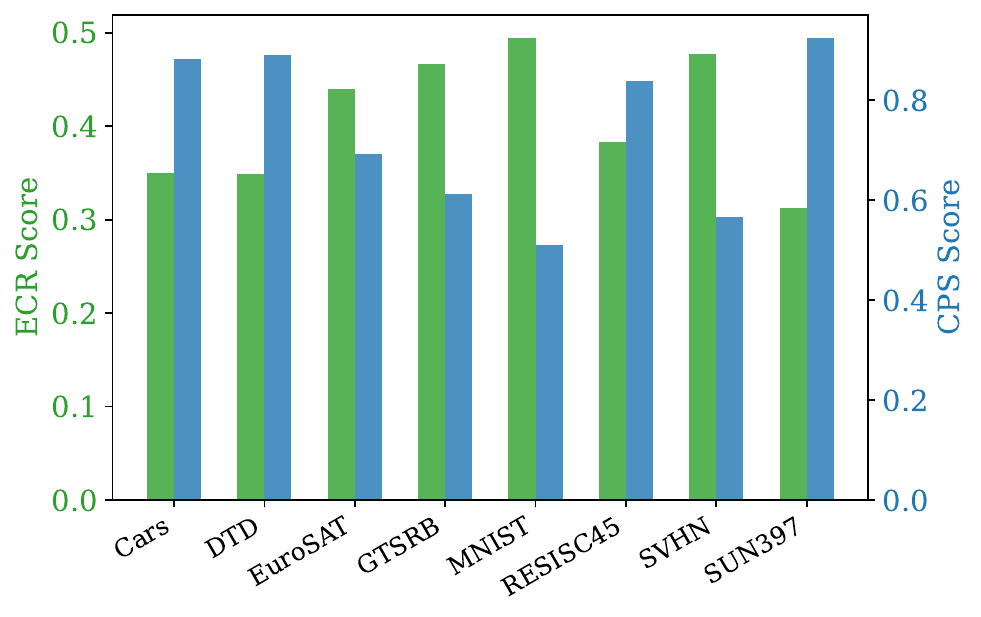}
    \caption{ViT-L-14, 8 tasks}
    \label{fig:rp_c1}
  \end{subfigure}
  \hfill
  \begin{subfigure}[b]{0.32\linewidth}
    \centering
    \includegraphics[width=\linewidth]{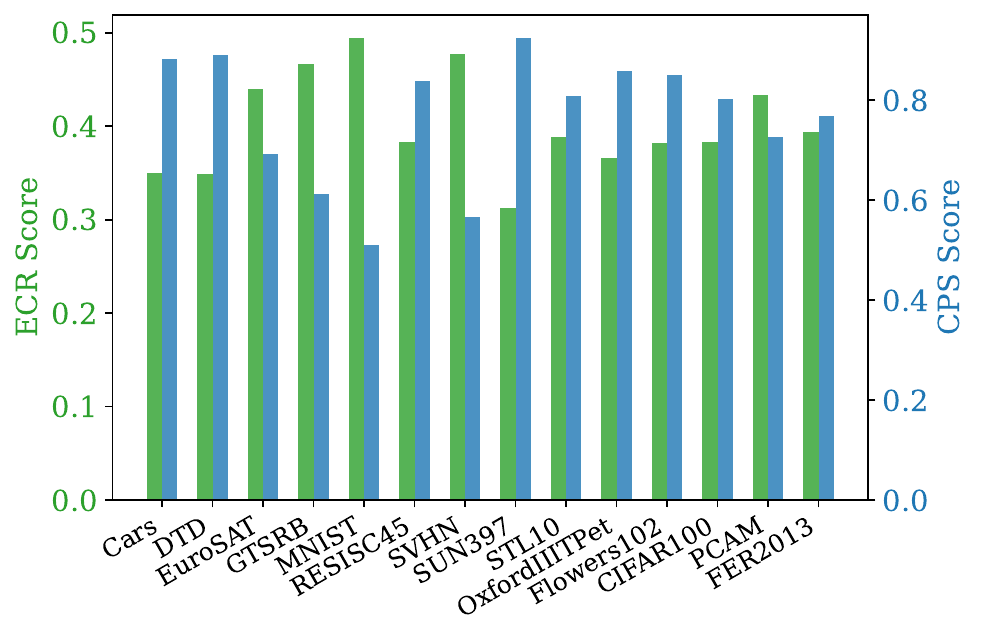}
    \caption{ViT-L-14, 14 tasks}
    \label{fig:rp_c2}
  \end{subfigure}
  \hfill
  \begin{subfigure}[b]{0.32\linewidth}
    \centering
    \includegraphics[width=\linewidth]{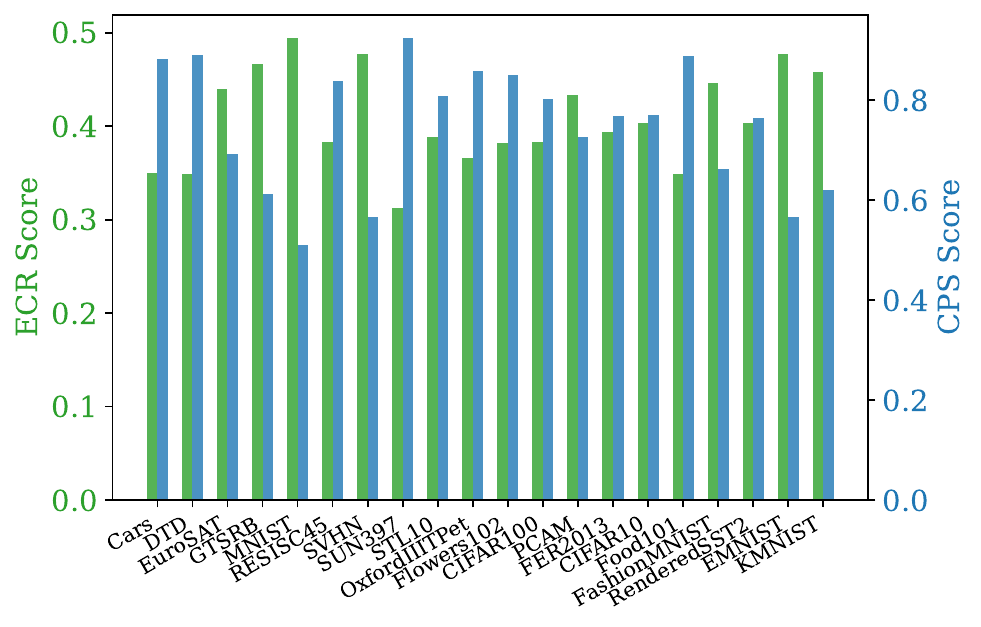}
    \caption{ViT-L-14, 20 tasks}
    \label{fig:rp_c3}
  \end{subfigure}

  \caption{Comparison of ECR and CPS metrics across different vision backbones.}
  \label{fig:rigidity_plasticity_all}
\end{figure*}

\begin{figure*}[htbp]
  \centering
  \includegraphics[width=1.0\linewidth]{figs/result_charts/FFT_vit32.pdf}
  \vspace{0.4cm}
  
  \includegraphics[width=1.0\linewidth]{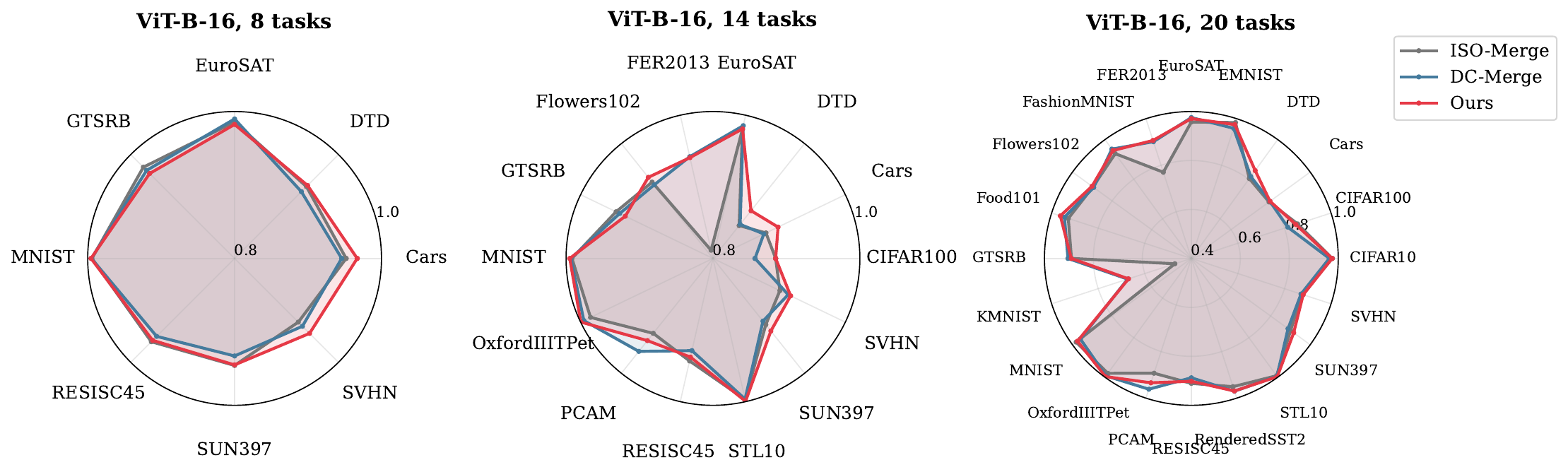}
  \vspace{0.4cm}
  
  \includegraphics[width=1.0\linewidth]{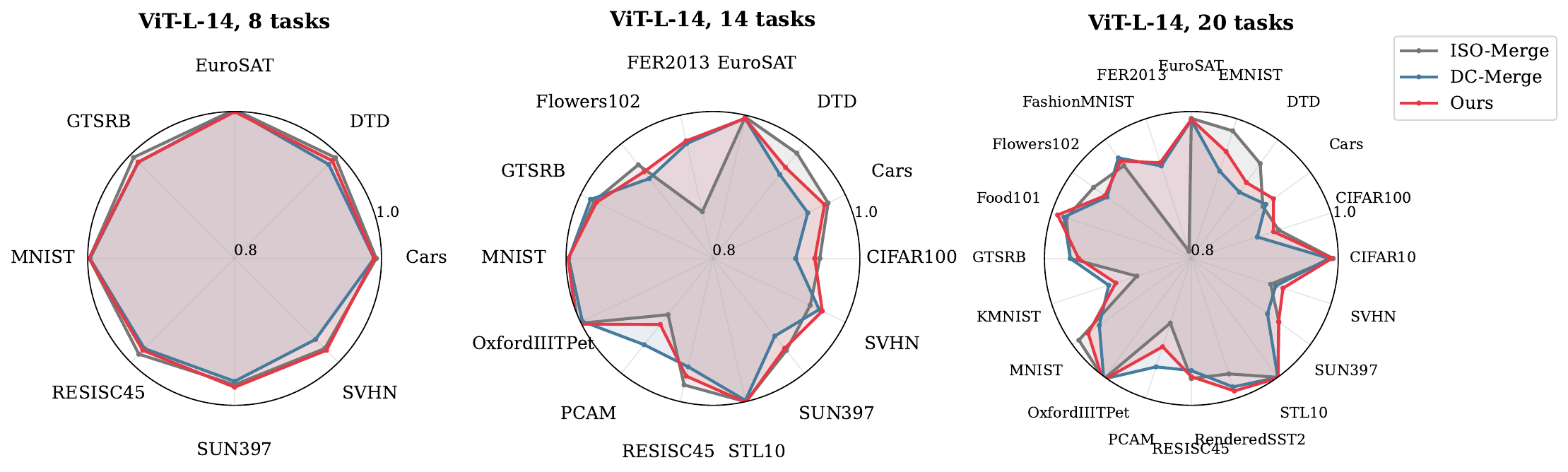}

  \caption{Radar‑chart performance comparison under the FFT setting for ViT‑B‑32, ViT‑B‑16 and ViT‑L‑14 backbones.}
  \label{fig:radar_fft_all}
\end{figure*}

\begin{figure*}[htbp]
  \centering
  \includegraphics[width=1.0\linewidth]{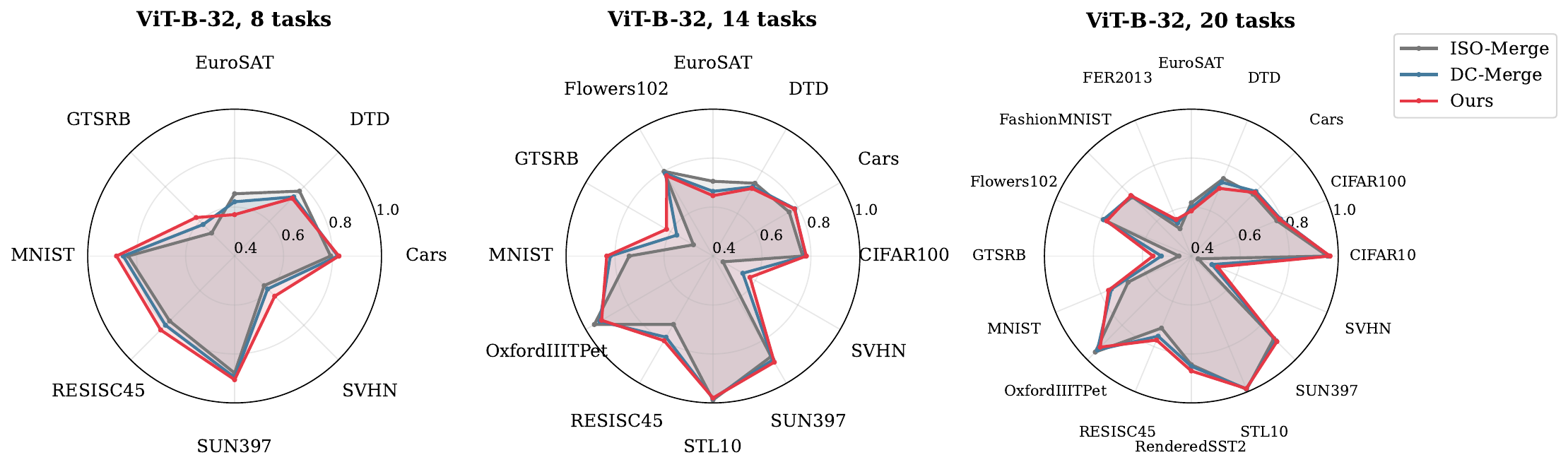}
  \vspace{0.4cm}

  \includegraphics[width=1.0\linewidth]{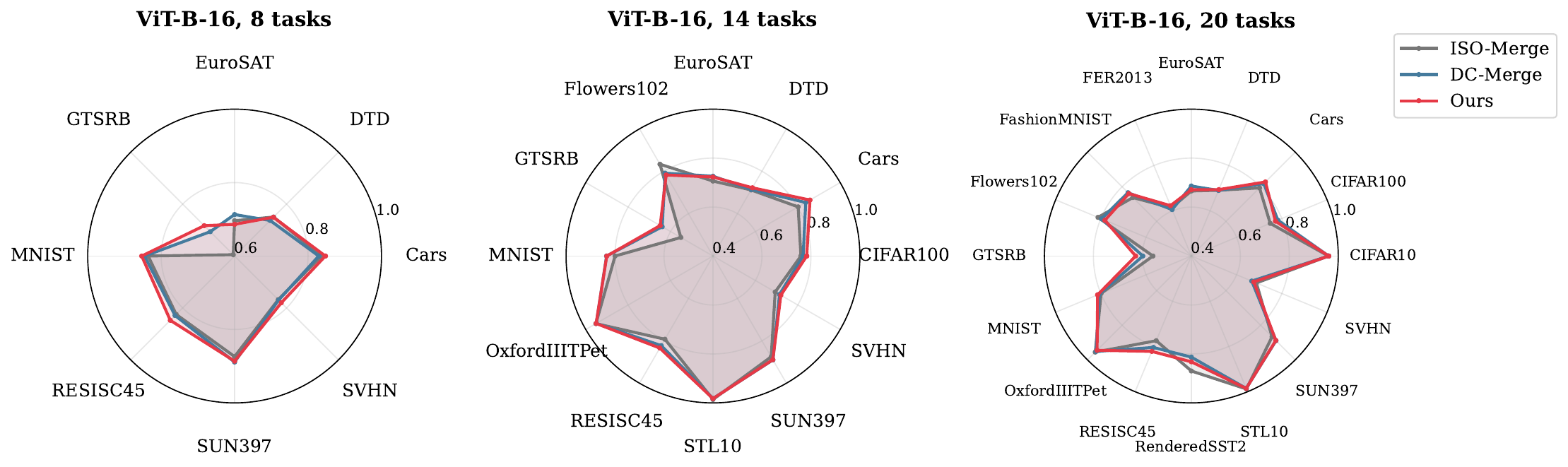}
  \vspace{0.4cm}

  \includegraphics[width=1.0\linewidth]{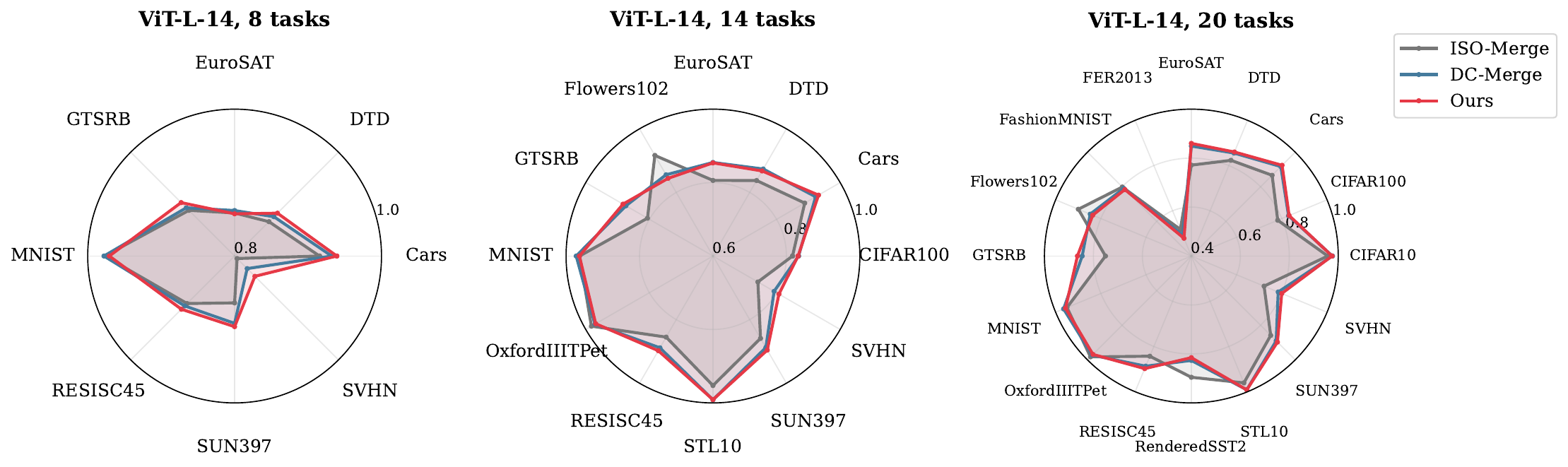}

  \caption{Radar‑chart performance comparison under the LoRA setting for ViT‑B‑32, ViT‑B‑16 and ViT‑L‑14 backbones.}
  \label{fig:radar_lora_all}
\end{figure*}

This section provides additional visualizations supporting the spectral analysis and experimental findings of SADA-Merging. Fig.~\ref{fig:sv_all_backbone} shows the singular-value distributions of task updates under the FFT setting, highlighting variations in spectral concentration across backbones and task scales. Fig.~\ref{fig:entropy_rank_all} illustrates the resulting spectral entropy-based rank allocation, demonstrating how subspace capacity varies across different settings. Fig.~\ref{fig:rigidity_plasticity_all} further visualizes the ECR and CPS metrics, characterizing task-wise spectral rigidity and plasticity. Finally, Fig.~\ref{fig:radar_fft_all} and Fig.~\ref{fig:radar_lora_all} present the complete radar-chart results under the FFT and LoRA settings, respectively, for ViT-B-32, ViT-B-16, and ViT-L-14 across different numbers of tasks.

\end{appendix}


\end{document}